\documentclass{gammarobot}

\usepackage[utf8]{inputenc}
\usepackage{amsmath,amssymb}

\usepackage{amsmath,amsfonts,bm}

\def\eqref#1{equation~\ref{#1}}

\def\1{\bm{1}}

\newcommand{\Tau}{\mathcal{T}}

\DeclareMathAlphabet{\mathsfit}{\encodingdefault}{\sfdefault}{m}{sl}
\SetMathAlphabet{\mathsfit}{bold}{\encodingdefault}{\sfdefault}{bx}{n}

\usepackage{colortbl}
\usepackage{wrapfig}
\usepackage{capt-of}
\usepackage{url}
\usepackage{fancyhdr}

\definecolor{PaperTheme}{HTML}{2D8FE3}
\colorlet{metablue}{PaperTheme!80!black}
\colorlet{metabg}{PaperTheme!8!white}
\colorlet{GammaPurple}{PaperTheme}
\colorlet{GammaCite}{metablue}
\colorlet{GammaFrame}{PaperTheme!60!white}
\colorlet{GammaBack}{PaperTheme!5!white}

\fancypagestyle{gammaTeleAI}{%
  \fancyhf{}
  \fancyfoot[C]{\thepage}

}
\makeatletter
\let\ps@plain\ps@gammaTeleAI
\makeatother

\renewcommand{\authorformat}[2][]{\mbox{\sffamily\bfseries #2$^{#1}$}}
\title{Test-Time Compute for Humanoid Behavior Foundation Models}
\author[1,2,\dagger]{Weiji Xie}
\author[1,2,\dagger]{Junsong Wu}
\author[1,2]{Zibo Zhou}
\author[1,2]{Yueru Chen}
\author[1,3]{Xingyi Wang}
\author[2]{Zhenhua Xiong}
\author[2]{Weinan Zhang}
\author[1,4,*]{Chenjia Bai}
\affiliation[1]{Institute of Artificial Intelligence (TeleAI), China Telecom}
\affiliation[2]{Shanghai Jiaotong University}
\affiliation[3]{Zhejiang University}
\affiliation[4]{Gamma Robotics}
\contribution{\textsuperscript{$\dagger$}Equal contribution\quad\textsuperscript{*}Corresponding author}
\hypersetup{
  pdftitle={Test-Time Compute for Humanoid Behavior Foundation Models},
  pdfauthor={Weiji Xie, Junsong Wu, Zibo Zhou, Yueru Chen, Xingyi Wang, Zhenhua Xiong, Weinan Zhang, Chenjia Bai}
}
\date{\today}
\project{\href{https://bfmttc.github.io/}{https://bfmttc.github.io/}}
\metadata[Correspondence to]{Chenjia Bai (\email{baicj@chinatelecom.cn})}
\abstract{\looseness=-1
Humanoid Behavior Foundation Models (BFMs) can execute impressive motions, but downstream applications such as loco-manipulation demand high whole-body positional accuracy. 
Existing BFMs suffer from a systematic command–execution gap, where executed motions deviate from desired ones. 
We show that additional test-time compute can improve execution without retraining the BFM. 
First, we introduce Test-Time Command Optimization (TTCO), which adapts the input command to the specific BFM to produce an executed trajectory closer to the desired motion. TTCO finds such commands by sampling-based optimization through closed-loop simulations of the BFM and robot dynamics.
Across three BFMs on the AMASS dataset, TTCO reduces mean global body-position error by 79–84\% in simulation.
Second, we analyze test-time compute scaling and derive an approximate compute-optimal strategy that can sustain performance gains as the compute budget increases.
Third, TTCO can generate BFM-specific action labels from recorded robot trajectories without requiring the original teleoperation body-motion commands. Simulation and real-robot experiments show that these labels are effective for training high-level loco-manipulation policies.
This method also supports reusing existing demonstrations across BFMs. 
}

\begin{document}
\addtocontents{toc}{\protect\setcounter{tocdepth}{-1}}
\maketitle

\par
\noindent\begin{minipage}{\linewidth}
    \centering
    \includegraphics[width=0.90\linewidth]{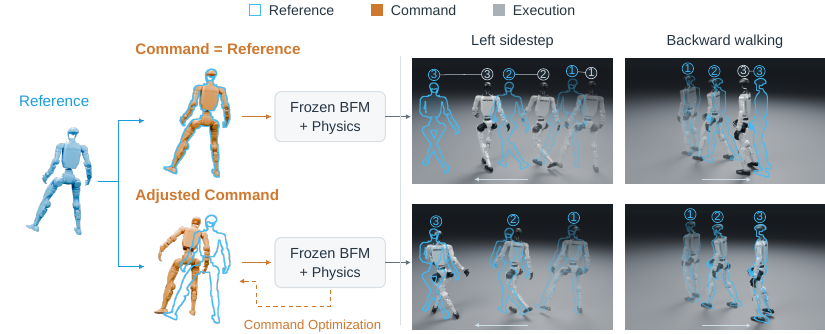}
    \setlength{\abovecaptionskip}{4pt}
    \captionof{figure}{\textbf{Changing the command can bring execution closer to the same reference.}
    Directly using the reference as the command can leave a command--execution gap.
    With the BFM fixed, our command optimization adjusts commands to bring execution closer to the same reference.
    The adjustment of commands is exaggerated for demonstration.}
    \label{fig:command-compensation}
\end{minipage}
\par\addvspace{\textfloatsep}

\section{Introduction}
\label{sec:introduction}

Recent advances in humanoid Behavior Foundation Models (BFMs) enable the
successful execution of diverse motions, including complex martial arts and
dance sequences~\citep{liao2025beyondmimic,luo2026sonic,xie2025kungfubot,ma2026rgmt,zeng2025bfm,wang2026omnixtreme,li2025bfmzero,zest,yuan2026surveybfm,yin2025unitracker}. These
models support downstream applications such as loco-manipulation
\citep{luo2026sonic,hu2026openhlm,wang2026vlk,ze2025twist2,xie2026grail,wu2026teleopit,chen2026holomotion}, which demand substantially higher whole-body
positional accuracy~\citep{jiang2025wholebodyvla}, ideally matching the accuracy of robotic arm
position controllers. Due to training data quality,
regularization design and other constraints~\citep{luo2026sonic}, current humanoid BFMs exhibit
systematic deviations between desired and executed motions, a discrepancy we
call the \emph{command--execution gap}. A robot can complete a movement while
its body drifts from the intended global position or its hands and feet miss the
intended contact locations. Prior work addresses these errors by
scaling training datasets, model size, and computational resources
\citep{luo2026sonic,qi2026humanoidgpt,zeng2026scalebfm}, as well as improving training procedures and network
architectures \citep{liu2026heft,wang2026omnixtreme,ma2026rgmt,han2026kungfubot2}.

\looseness=-1
\textbf{We show that additional test-time compute can improve motion
execution without changing the BFM.} At inference time, the desired motion
need not be supplied directly as the command to the BFM. Adapting the input
command to the specific BFM can yield an executed trajectory that more
closely matches the original desired motion
(Figure~\ref{fig:command-compensation}).
We introduce \emph{Test-Time Command Optimization} (TTCO), which finds such
commands through sampling-based optimization with closed-loop simulations
of the BFM and robot dynamics.
We evaluate TTCO on 12,273 AMASS~\citep{mahmood2019amass} reference motions
using HEFT~\citep{liu2026heft}, SONIC~\citep{luo2026sonic}, and
TWIST2~\citep{ze2025twist2}.
Compared with directly using the reference motion as the command, TTCO
reduces mean global body-position error by 79--84\% across these three
BFMs in simulation.

\noindent\begin{minipage}[t]{0.60\linewidth}
\vspace{0pt}
This framework provides a direct way to scale up test-time compute. The gains from command optimization depend on both the amount of compute and its allocation across candidates, optimization iterations, rollout horizon, and replanning. This raises two questions:
\emph{To what extent can additional compute improve performance, and
how should a fixed compute budget be allocated to achieve the best performance?}
To answer these questions, we analyze the compute-allocation trade-offs and construct a surrogate cost model fitted to limited measurements on a small motion set. This model yields an approximate compute-optimal strategy that sustains performance gains as compute increases (Figure~\ref{fig:compute-strategy}) and generalizes beyond the motions and BFM used for fitting.
\end{minipage}\hfill
\begin{minipage}[t]{0.38\linewidth}
    \vspace{0pt}
    \centering
    \setlength{\abovecaptionskip}{3pt}
    \includegraphics[width=\linewidth]{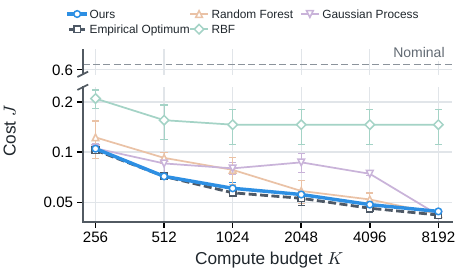}
    \small
    \captionof{figure}{\textbf{Scaling test-time compute with an approximate
    compute-optimal strategy.}
    Our proposed strategy achieves the best overall performance among four strategies.}
    \label{fig:compute-strategy}
\end{minipage}
\par

\looseness=-1
The benefits of command optimization extend to downstream humanoid
loco-manipulation. Because of the command--execution gap, a recorded robot
trajectory generally cannot serve as an action label for high-level policy learning. Existing pipelines therefore train high-level policies
with teleoperation commands tied to the BFM used for demonstration collection~\citep{hu2026openhlm,ze2025twist2,luo2026sonic}. In
contrast, we apply TTCO offline to optimize body-motion commands against the recorded
trajectories while keeping the target BFM fixed. Experiments in HumanoidArena~\citep{wang2026humanoidarena} and on real robots show that optimized commands provide effective supervision for high-level policy learning, reducing dependence on the original body-motion labels and supporting demonstration reuse across BFMs.

In summary, our contributions to test-time compute for humanoid BFMs are:

\begin{list}{\textbullet}{\setlength{\leftmargin}{1em}\setlength{\labelsep}{0.5em}\setlength{\topsep}{0pt}\setlength{\partopsep}{0pt}\setlength{\itemsep}{0pt}\setlength{\parsep}{0pt}}
\item \textbf{We introduce TTCO to improve motion-tracking fidelity in existing humanoid BFMs through test-time compute, without further training.} Across three pretrained BFMs and 12,273 AMASS reference motions, TTCO reduces mean global body-position error by 79--84\% in simulation.

\item \textbf{We analyze test-time compute scaling and derive an approximate compute-optimal allocation strategy.} Our surrogate cost model captures allocation trade-offs and yields a strategy that generalizes across motions and BFMs.

\item \textbf{We extend test-time compute to high-level policy learning and demonstration reuse across BFMs.} We validate TTCO-generated body-action labels for both uses in simulation and on real robots.
\end{list}

\section{Related Work}
\label{sec:related-work}

\paragraph{Humanoid Behavior Foundation Models.}
Humanoid BFMs build on advances in whole-body motion tracking, enabling the execution of diverse and complex skills~\citep{liao2025beyondmimic,han2026kungfubot2,ma2026rgmt,zhu2026clot,li2026omniclone,wang2026omnixtreme}.
These models provide a reusable motion-command
interface~\citep{zeng2025bfm,zeng2026scalebfm,li2025bfmzero,luo2026sonic}, through which high-level policies \citep{ze2025twist2,hu2026openhlm} can specify target motions to perform downstream loco-manipulation.
Recent work scales these BFMs at training time through more compute, data and capacity \citep{luo2026sonic,qi2026humanoidgpt,chen2026holomotion,zeng2026scalebfm}.
We study inference compute for humanoid BFMs in general and how tracking quality benefits from scaling that compute.

\paragraph{Sampling-Based Methods for Robot Control.}
Sampling-based methods optimize control sequences through physics rollouts without requiring derivatives of the dynamics or cost~\citep{williams2018mppi,howell2022predictive,xue2025dialmpc,liu2026silr,pan2025spider,dhedin2026dynaretarget,jordana2025zeroorder}.
Sumo~\citep{zhang2026sumo} applies this approach in the command space of a pretrained whole-body policy, using policy-in-the-loop simulations to plan object interactions.
BFM-Zero~\citep{li2025bfmzero} also optimizes latent prompt sequences in simulation to improve tracking, with discussion limited to its own model.
We apply sampling-based optimization to pretrained humanoid BFMs in general, to optimize motion commands and reduce execution error relative to a fixed reference motion.

\paragraph{Test-Time Compute Scaling and Allocation.}
\looseness=-1
Language-model studies show that test-time performance depends on both the compute budget and its allocation to sampling, refinement, and verification~\citep{brown2024monkeys,zhang2025surveytts,snell2025scaling,wu2025inference,setlur2025verification}.
Embodied systems use test-time compute to sample and select actions \citep{kwok2025robomonkey} or executable motions \citep{cao2026texedo}, and some adaptively allocate that compute between action refinement and parallel sampling in a frozen generative policy \citep{li2026elastic}.
We measure test-time compute by the number of simulated BFM transitions, then study the scaling and allocation of that budget for pretrained humanoid BFMs.

\section{Method}
\label{sec:method}

\subsection{Problem Setting}
\label{sec:method-preliminaries}

\looseness=-1
\noindent\textbf{Behavior Foundation Model.}\ We consider a humanoid Behavior Foundation
Model (BFM) $\pi$ controlling a robot to track a reference motion.  Let $\mathcal M_{1:T}$ denote the motion-sequence space over
time steps $1:T$.  Let
$\tau^{\mathrm{ref}}_{1:T}\in\mathcal M_{1:T}$ be the desired reference motion,
where $\tau_t^{\mathrm{ref}}\in\mathbb R^{2n_q+13}$ represents the robot's
generalized coordinates and velocities at time $t$, and $n_q$ is the number of
joints.  Let $s_t$, $o_t$, and $a_t$ be the humanoid state, policy observation,
and action. The BFM $\pi$ receives a \emph{motion command} sequence
$c_{1:T}\in\mathcal M_{1:T}$ as its task input and produces action $a_t$ based on
observation history $o_{0:t-1}$ at each time step. We assume that both the
policy $\pi$ and the environment dynamics $f$ are deterministic and the
initial state is fixed to $s_0$. Then we have
$a_t=\pi(o_{0:t-1},c_t)$, $(s_t,o_t)=f(s_{t-1},a_t)$.\footnote{While some BFMs also use past and future command frames, we consider only the current frame for notational simplicity.}
With $z_t=(s_t,o_{0:t})$, the BFM policy and environment dynamics jointly
define the closed-loop transition $z_t=F_\pi(z_{t-1},c_t)$. For the fixed
initial state, we use $\Tau_\pi(c)$ to denote the executed trajectory induced
by the motion command $c$.
Appendix~\ref{app:general-motion-tracking} describes the motion-tracking
formulation, training, and control interface underlying this abstraction.

\noindent\textbf{Command-Execution Gap.}\ Standard BFM inference directly uses the desired reference motion as the motion
command input~\citep{luo2026sonic,ze2025twist2}.  We call it the \emph{nominal motion command},
$c^{\mathrm{nom}}=\tau^{\mathrm{ref}}_{1:T}$, which produces the executed
trajectory 
$\tau^{\mathrm{nom}}=\Tau_\pi(c^{\mathrm{nom}})$.
In general, a BFM is trained to minimize the tracking loss between the executed
and reference trajectories.
However, a BFM cannot execute $\tau^{\mathrm{ref}}$ exactly because reference
motions may be physically infeasible, the training motion data may have
limited coverage, policy optimization may leave residual errors, and the
regularization and domain randomization used for sim-to-real transfer may
trade tracking precision for robustness.  Consequently, the executed
trajectory $\tau^{\mathrm{nom}}$ systematically deviates from the desired reference
motion $\tau^{\mathrm{ref}}$, creating a \emph{command--execution gap}.

\looseness=-1
\noindent\textbf{Command Optimization.}\ To address this gap, our key insight is that \emph{the motion command that
produces the most accurate execution need not equal the desired reference
motion}.  Compensating for the systematic deviation of a BFM, we can find an alternative
command $c^\star$ that deviates from $c^{\mathrm{nom}}$ and produces an
executed trajectory $\Tau_\pi(c^\star)$ that matches
$\tau^{\mathrm{ref}}$ more closely than $\tau^{\mathrm{nom}}=\Tau_\pi(c^{\mathrm{nom}})$.

Formally, we measure the tracking error of a candidate motion command $c$ as
$\mathcal{L}^{\mathrm{track}}(c,\tau^{\mathrm{ref}})
=\sum_{t=1}^{T}\ell_t^{\mathrm{track}}
(\Tau_\pi(c)_t,\tau_t^{\mathrm{ref}})$.
To penalize excessive command changes and jitter, we add a regularization term
$\mathcal{L}^{\mathrm{reg}}(c,\tau^{\mathrm{ref}})
=\sum_{t=1}^{T}\ell_t^{\mathrm{reg}}
(c_t,\Tau_\pi(c)_t,\tau_t^{\mathrm{ref}})$ and define the cost function
as $J(c,\tau^{\mathrm{ref}})
=\mathcal{L}^{\mathrm{track}}(c,\tau^{\mathrm{ref}})
+\mathcal{L}^{\mathrm{reg}}(c,\tau^{\mathrm{ref}})$.
The optimal motion command is then given by
\begin{equation}
    c^\star=\operatorname*{argmin}_{c\in\mathcal M_{1:T}}
    J(c,\tau^{\mathrm{ref}}).
    \label{eq:command-optimization-problem}
\end{equation}
We refer to this optimization as the \emph{full-sequence command-optimization
problem}.  Exact definitions of all loss terms are provided in
Appendix~\ref{app:rollout-cost}.

\subsection{Test-Time Command Optimization}
\label{sec:test-time-command-optimization}

\begin{figure}[t]
    \centering
    \setlength{\abovecaptionskip}{4pt}
    \includegraphics[width=0.80\linewidth]{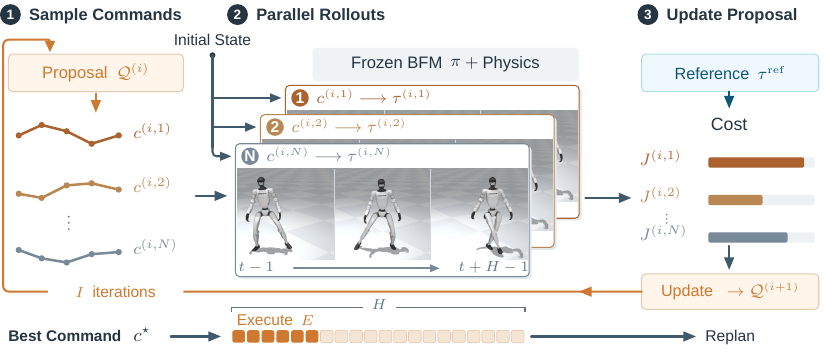}
    \caption{\textbf{Overview of test-time command optimization.} We evaluate $N$ candidates through parallel $H$-step rollouts from the same state and update the proposal using their local costs. After $I$ iterations, we execute the first $E$ steps of the lowest-cost candidate in simulation, then replan.}
    \label{fig:test-time-command-optimization}
\end{figure}

\noindent\textbf{Receding-Horizon Command Optimization.}\ Direct optimization of the
non-differentiable, long-horizon problem is intractable: the number of
decision variables grows with $T$, and each candidate evaluation requires
$T$ simulated BFM transitions. We therefore
approximate $c^\star$ using receding-horizon optimal control~\citep{rawlings2017mpc}.  For a length-$T$
reference motion, we solve a sequence of finite-horizon local problems: at each
planning step, we optimize an $H$-step motion-command segment, execute its first
$E$ steps in simulation, and formulate the next local problem from the
resulting state (Figure~\ref{fig:test-time-command-optimization}).

At time $t$, the local problem is formulated as follows:
\begin{align}
    c^\star_{t:t+H-1}
    &=\operatorname*{argmin}_{c_{t:t+H-1}\in\mathcal M_{t:t+H-1}}
      J_{\mathrm{local}}(c_{t:t+H-1},z_{t-1},\tau^{\mathrm{ref}}_{t:t+H-1}),
    \label{eq:finite-horizon-problem}\\
    J_{\mathrm{local}}
    &=\sum_{h=0}^{H-1}
      \left(\ell_{t+h}^{\mathrm{track}}+\ell_{t+h}^{\mathrm{reg}}\right)
      +\ell^{\mathrm{term}}_{t+H-1},
    \label{eq:finite-horizon-cost}\\
    \text{s.t.}\qquad z_k
    &=F_\pi(z_{k-1},c_k),
    \qquad k=t,\ldots,t+H-1.
    \label{eq:closed-loop-forward-dynamics}
\end{align}
\looseness=-1
The terminal cost $\ell_{t+H-1}^{\mathrm{term}}$ includes the joint- and
body-position tracking errors at the end of the planning horizon.  It serves
as a surrogate that accounts for the effect of costs beyond the finite planning
horizon.  For notational simplicity, we omit the arguments of the cost terms
when they are clear from context.

\noindent\textbf{Sampling-based Optimization.}\ We solve each local problem with a sampling-based optimizer~\citep{jordana2025zeroorder}. We evaluate each candidate command by running the BFM in a physics simulator. To simplify the optimization, we define the decision vector
$x$ as the command values at control knots spaced $H_k$ steps apart and
reconstruct the complete motion-command segment through interpolation,
$c=\Phi(x)$~\citep{howell2022predictive}, with details in
Appendix~\ref{app:command-implementation}.

The optimizer performs $I$ iterations of the following three steps.
(1)~Sampling: At iteration $i$, it samples $N$ candidates
$x^{(i,n)}\sim\mathcal Q^{(i)}$, $n=1,\ldots,N$, from the current proposal
distribution $\mathcal Q^{(i)}$.  (2)~Rollout: Each candidate is interpolated
into $c^{(i,n)}=\Phi(x^{(i,n)})$ and rolled out from $z_{t-1}$ through the
BFM and physics simulator to evaluate its local cost
$J^{(i,n)}=J_{\mathrm{local}}(c^{(i,n)},z_{t-1},
\tau^{\mathrm{ref}}_{t:t+H-1})$.  (3)~Update: The resulting candidate--cost pairs
are used to update the proposal distribution for the next iteration,
$\mathcal Q^{(i+1)}=\operatorname{Update}(\mathcal Q^{(i)},
\{x^{(i,n)},J^{(i,n)}\}_{n=1}^{N})$.
After $I$ iterations, we execute the first $E$ steps of the lowest-cost
candidate found across these iterations. 

\looseness=-1
We consider five sampling-based optimizers: Cross-Entropy Method (CEM)~\citep{deboer2005cem},
Predictive Sampling (PS)~\citep{howell2022predictive}, Model Predictive Path Integral with covariance
matrix adaptation (MPPI-CMA)~\citep{jordana2025zeroorder}, CMA-ES~\citep{hansen2001cmaes}, and
Consensus-based Optimization (CBO)~\citep{sun2026cbo}.
Based on the analysis in Secs.~\ref{sec:test-time-compute} and~\ref{sec:compute-configuration-effects}, CEM achieves competitive performance overall and is selected for the subsequent analysis and experiments.

\subsection{Scaling Test-Time Compute Optimally}
\label{sec:test-time-compute}

\noindent\textbf{Test-Time Compute-Optimal Strategy.}\ Finite horizons and
finite sampling make the problem tractable, but can increase execution cost relative to
the full-sequence optimum. 
Longer horizons($H$) account for more future cost, more frequent replanning(smaller $E$) revises the plan earlier, and more iterations($I$) or candidates($N$) can improve local search. These choices compete for a given compute budget.
We therefore seek the algorithm
and compute configuration that minimize full-sequence execution cost at each budget.
For compute configuration $\xi=(I,N,H,E)\in\mathcal X$, where
$\mathcal X=\{(I,N,H,E)\in\mathbb N_+^4:1\le E\le H\}$,
a length-$T$ motion requires $\lceil T/E\rceil$ planning calls, each using
$INH$ simulated BFM transitions to evaluate candidates. Thus
$C(\xi,T)=\lceil T/E\rceil INH$ measures the total compute. To compare compute across motion lengths,
we define the \emph{compute budget ratio} $K(\xi)=INH/E$,
so $C(\xi,T)\approx TK(\xi)$.

Let $J_{\mathrm{TTC}}(\mathcal A,\xi)$ denote the expected full-sequence
cost $J$ per time step, averaged over the motion dataset, when using
algorithm $\mathcal A$ and configuration $\xi$.
The compute-optimal strategy maps each budget $K$ to an algorithm and a
configuration that solve
\begin{equation}
    \begin{aligned}
        (\mathcal A^\star(K),\xi^\star(K))
        &=\operatorname*{argmin}_{\mathcal A,\,\xi\in\mathcal X:\,K(\xi)\le K}
        J_{\mathrm{TTC}}(\mathcal A,\xi).
    \end{aligned}
    \label{eq:compute-optimal-configuration}
\end{equation}
We call $\xi^\star(K)$ the \emph{compute-optimal configuration}.

\noindent\textbf{Modeling Test-Time Performance.}\ An exhaustive search over
$(\mathcal A,\xi)$ is computationally prohibitive, since evaluating
$J_{\mathrm{TTC}}$ for each configuration requires command optimization
over the motion dataset. We therefore construct a
heuristic \emph{surrogate cost model} to predict costs for untested
configurations from limited measurements. Its structure follows
the finite-horizon and sampling approximations in
Sec.~\ref{sec:test-time-command-optimization}.
We represent the irreducible cost of the full-sequence problem by $J_\infty$.
The finite-horizon error $\epsilon_{\mathrm{horizon}}$ can remain with exact
local solutions; we model it as independent of $(\mathcal A,I,N)$.
Finite local search adds the sampling error $\epsilon_{\mathrm{sample}}$:
\begin{equation}
    \widehat J_{\mathrm{TTC}}(\mathcal A,\xi)
    =J_\infty+\epsilon_{\mathrm{horizon}}(H,E)
     +\epsilon_{\mathrm{sample}}(\mathcal A,I,N\mid H,E).
    \label{eq:cost-model-decomposition}
\end{equation}
The terminal cost in $J_{\mathrm{local}}$ approximates costs beyond $H$.
\citet{kunisch2020terminal} prove that, in their
stabilization setting, the cost gap caused by this approximation has an
upper bound that decays exponentially with $2H-E$ (in our notation).
This motivates an exponential term that decreases with longer planning
horizons and shorter execution intervals.
Executing more steps before replanning can also reduce robustness to
disturbances \citep{gruene2015robustness}. We account for this dependence
with an empirical linear correction, giving the finite-horizon error model
\begin{equation}
    \epsilon_{\mathrm{horizon}}(H,E)
    =C_H\exp[-\alpha_H(2H-E)]+C_E(E-1).
    \label{eq:execution-cost-model}
\end{equation}
\looseness=-1
To model the sampling error, we first isolate the effect of solver effort
$(I,N)$ at fixed $(H,E)$. Iterations refine the proposal distribution;
candidates explore the command space within each update. At a fixed $IN$, more iterations
leave fewer candidates per update. We represent this dependence by the
solver-dependent error $\epsilon_{\mathrm{solver}}$, using additive
inverse-power terms to capture diminishing returns
\citep{hoffmann2022computeoptimal}:
\begin{equation}
    \epsilon_{\mathrm{solver}}(\mathcal A,I,N)
    =C_I(\mathcal A)I^{-\alpha_I(\mathcal A)}
     +C_N(\mathcal A)N^{-\alpha_N(\mathcal A)}.
    \label{eq:solver-cost-model}
\end{equation}
We next account for the effect of $H$ on sampling error. With control-knot
spacing $H_k$ fixed, the dimension $d$ of the decision vector $x$ grows
approximately in proportion to $H$. For Lipschitz optimization with a
fixed Lipschitz constant, the minimax rate $n^{-1/d}$ for $n$ evaluations
motivates raising $\epsilon_{\mathrm{solver}}$ to $\eta/H$
\citep{pmlr-v70-malherbe17a}. This makes improvement with solver effort
slower at longer horizons. A separate factor $\gamma^H$ allows the error
scale to vary with $H$, giving
\begin{equation}
    \epsilon_{\mathrm{sample}}(\mathcal A,I,N\mid H,E)
    =\epsilon_{\mathrm{solver}}(\mathcal A,I,N)^{\eta/H}\gamma^H.
    \label{eq:optimization-cost-model}
\end{equation}
With positive parameters and fixed $(H,E)$, the sampling error vanishes
as $I,N\to\infty$, leaving $J_\infty+\epsilon_{\mathrm{horizon}}$.

With measurements from different configurations, we can fit the surrogate cost model by minimizing the mean squared
difference between predicted and measured log costs. Minimizing the fitted model
under the same compute budget approximately solves
Eq.~\ref{eq:compute-optimal-configuration}:
\begin{equation}
    (\widehat{\mathcal A}^\star(K),\widehat{\xi}^\star(K))
    =\operatorname*{argmin}_{\mathcal A,\,\xi\in\mathcal X:\,K(\xi)\le K}
    \widehat J_{\mathrm{TTC}}(\mathcal A,\xi).
    \label{eq:predicted-compute-optimal-strategy}
\end{equation}
In our experiments, we restrict configurations to the practical search range
specified in Appendix~\ref{app:strategy-budget-protocol}.

\section{Experiments}
\label{sec:experiments}

\begin{table}[t]
    \centering
    \setlength{\abovecaptionskip}{4pt}
    \caption{\textbf{TTCO reduces mean cost and global
    position error at the price of higher joint acceleration, across three BFMs on AMASS.} Each entry reports nominal
    $\rightarrow$ optimized performance, averaged over all motions per BFM.}
    \label{tab:exp-amass}
    \small
    \setlength{\tabcolsep}{3.5pt}
    \resizebox{\linewidth}{!}{%
    \begin{tabular}{@{}lccccc@{}}
        \toprule
        BFM & \multicolumn{1}{c}{Mean cost $\downarrow$}
        & \multicolumn{1}{c}{Cost-win rate (\%) $\uparrow$}
        & \multicolumn{1}{c}{Body-position error (m) $\downarrow$}
        & \multicolumn{1}{c}{Root-position error (m) $\downarrow$}
        & \multicolumn{1}{c}{Joint acceleration (rad/s$^2$) $\downarrow$} \\
        \midrule
        HEFT & $0.419\!\rightarrow\!0.046$ & 62.27
        & $0.271\!\rightarrow\!0.057$
        & $0.264\!\rightarrow\!0.048$
        & $10.104\!\rightarrow\!17.246$ \\
        SONIC & $6.047\!\rightarrow\!0.323$ & 86.00
        & $0.615\!\rightarrow\!0.127$
        & $0.609\!\rightarrow\!0.117$
        & $\phantom{0}8.832\!\rightarrow\!11.999$ \\
        TWIST2 & $5.321\!\rightarrow\!0.573$ & 99.67
        & $1.095\!\rightarrow\!0.173$
        & $1.079\!\rightarrow\!0.164$
        & $43.280\!\rightarrow\!51.574$ \\
        \bottomrule
    \end{tabular}}
\end{table}

\begin{figure}[!b]
    \centering
    \begin{minipage}[c]{0.52\linewidth}
        \caption{\textbf{Real-robot demonstration of motion tracking with nominal and optimized commands.} We compare nominal and optimized commands for left sidestep and backward walking executed open-loop on the real robot. In both examples, optimized commands bring the final displayed positions closer to the reference. Blue outlines show the reference motion.}
        \label{fig:tracking-demo}
    \end{minipage}\hfill
    \begin{minipage}[c]{0.45\linewidth}
        \centering
        \includegraphics[width=\linewidth]{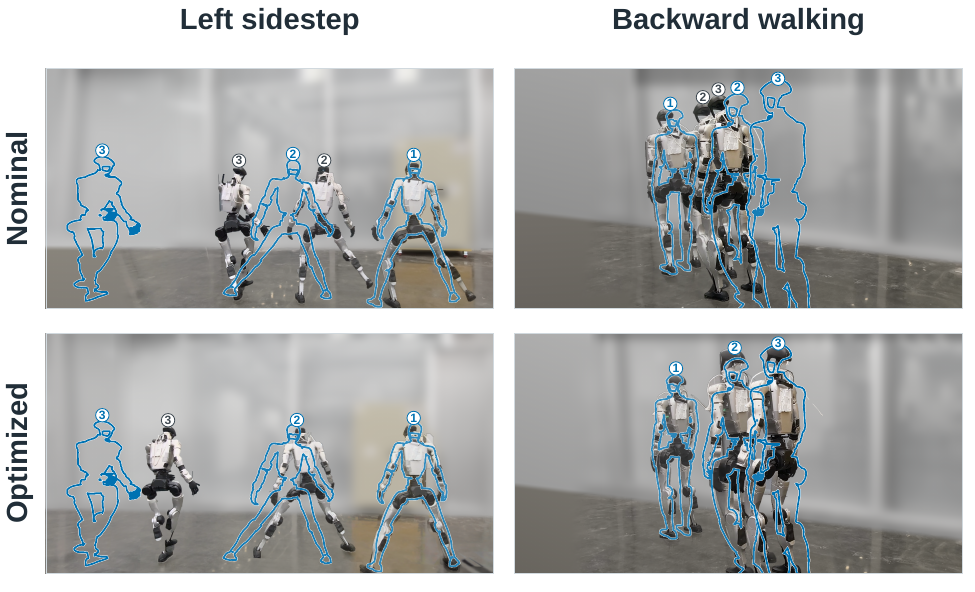}
    \end{minipage}
\end{figure}

We address four questions:

\begin{list}{\textbullet}{\setlength{\leftmargin}{1em}\setlength{\labelsep}{0.5em}\setlength{\topsep}{0pt}\setlength{\partopsep}{0pt}\setlength{\itemsep}{0pt}\setlength{\parsep}{0pt}}
\item \textbf{Q1:} Does command optimization improve tracking across pretrained BFMs at scale?

\item \textbf{Q2:} How does the allocation of test-time compute affect tracking performance?

\item \textbf{Q3:} Can the cost model find compute-optimal configurations across budgets, BFMs, and motions?

\item \textbf{Q4:} Does command optimization improve downstream loco-manipulation policies?
\end{list}

\noindent\textbf{Experimental setup.}\ We evaluate SONIC~\citep{luo2026sonic}, HEFT~\citep{liu2026heft}, and TWIST2~\citep{ze2025twist2} on G1. MuJoCo Warp~\citep{mujoco,mujoco_warp} supports tracking and command optimization; simulation for loco-manipulation uses Isaac Sim.

\subsection{Test-Time Command Optimization Improves Tracking Precision (Q1)}
\label{sec:large-scale-tracking}

\looseness=-1
For each pretrained BFM, we compare tracking performance with nominal and
optimized commands on the retargeted AMASS motion dataset~\citep{mahmood2019amass}. Appendix~\ref{app:large-scale-motion-tracking}
provides the evaluation protocol and metric definitions.

Command optimization reduces mean cost by 89.0--94.7\% and mean
global body-position error by 79.1--84.2\% across all three BFMs
(Table~\ref{tab:exp-amass}), with cost-win rates of 99.67\% for TWIST2 and
86.00\% for SONIC. Joint acceleration increases for all three BFMs,
indicating a trade-off between tracking accuracy and motion smoothness.
Note that we use a moderate compute budget ratio of K=640 to limit the cost of evaluating the full dataset; the results in Sections 4.2 and 4.3 suggest that larger budgets may further improve performance.
As shown in the two selected real-robot sequences in Figure~\ref{fig:tracking-demo},
optimized commands bring the resulting robot positions closer to the
reference.

\subsection{Compute Allocation Governs the Gains (Q2)}
\label{sec:compute-configuration-effects}

We examine the trade-offs that govern compute allocation by separately
varying solver parameters $(\mathcal A,I,N)$ and time horizon parameters
$(H,E)$, following the error decomposition in
Sec.~\ref{sec:test-time-compute}. Both sweeps use SONIC on the same
10 randomly sampled AMASS motions. The solver sweep fixes $(H,E)=(40,20)$
and compares algorithms and iteration--candidate splits at each budget
$O=IN$. To study the horizon error in isolation, we vary $(H,E)$ and
run CEM to convergence with $N=512$, approximately treating the remaining solver error as
negligible ($\epsilon_{\mathrm{solver}}\approx 0$).

\begin{figure}[!t]
    \centering
    \setlength{\abovecaptionskip}{4pt}
    \resizebox{\linewidth}{!}{\includegraphics{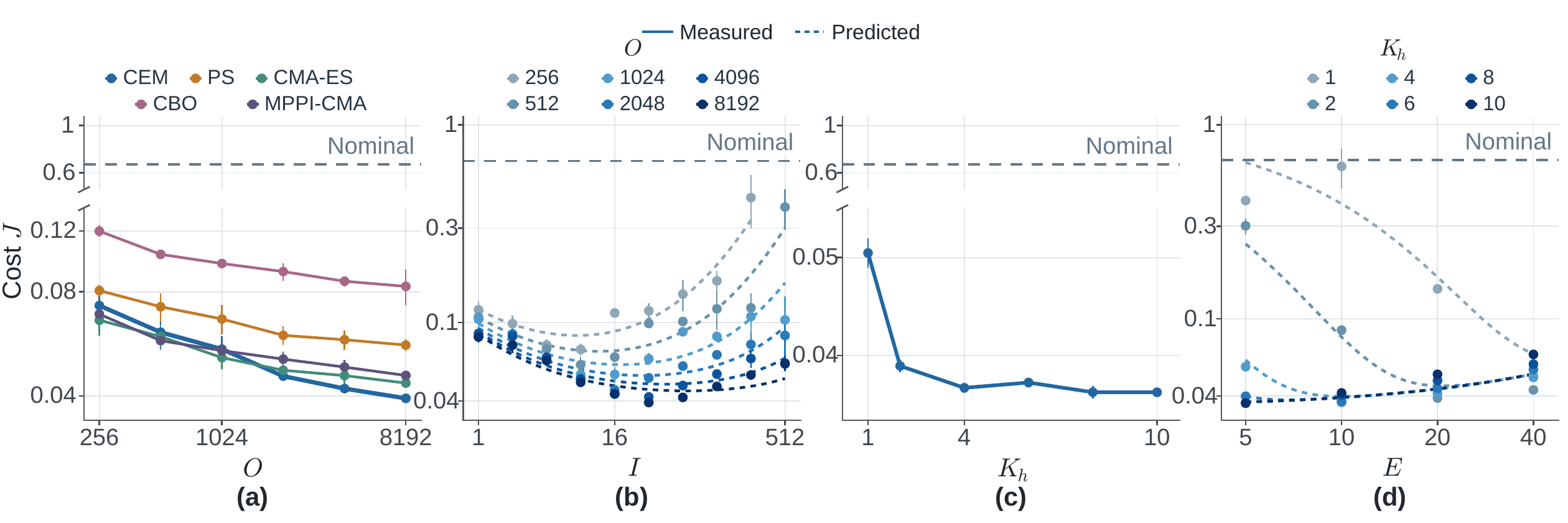}}
    \caption{\textbf{Compute allocation governs performance through solver and time horizon parameters.}
    We examine these two parameter groups separately.
    \textbf{(a,b) Solver parameters $(\mathcal A,I,N)$.} (a)
    Each point reports the lowest measured cost over all tested $(I,N)$
    configurations for that algorithm at that budget.
    CEM achieves competitive performance overall, so we restrict the remaining analysis to it.
    (b) At fixed $O=IN$, its cost first falls and then rises as iterations increase
    and candidates per iteration decrease; the best measured configuration
    changes with the budget.
    \textbf{(c,d) Time horizon parameters $(H,E)$.}
    (c) The best measured cost over $E$ levels off beyond a horizon ratio
    $K_h=H/E$ of 4. (d) Increasing $E$ can either lower or raise cost depending on $K_h$, motivating joint selection of $H$ and $E$.
    In (b,d), model curves follow the measurements, supporting the proposed
    forms of the solver and horizon error terms.
    Error bars show the standard deviation of the three seed means;
    horizontal dashed lines mark nominal-command cost.}
    \label{fig:compute-allocation}
\end{figure}

\noindent\textbf{Solver parameters $(\mathcal A,I,N)$ (Fig.~\ref{fig:compute-allocation}(a,b)).}\ CEM achieves the lowest cost at the three largest
budgets and remains competitive at the three smaller ones, so we restrict the
remaining analysis to CEM. At a fixed solver budget $O=IN$, cost first
decreases and then increases as more iterations leave fewer candidates per
iteration. For example, at $O=8192$, changing $(I,N)$ from $(1,8192)$ to
$(32,256)$ reduces cost by 53.42\% without increasing compute. The best
measured configuration also varies with the budget. The surrogate model captures this
trade-off and fits costs with a mean absolute relative error of 11.38\%
(Fig.~\ref{fig:compute-allocation}(b)), supporting our model of solver error.

\looseness=-1
\noindent\textbf{Time horizon parameters $(H,E)$ (Fig.~\ref{fig:compute-allocation}(c,d)).}\ The lowest measured cost across the tested $E$ values levels off beyond $K_h=4$ (Fig.~\ref{fig:compute-allocation}(c)).
The best measured configuration tends toward more
frequent replanning as $K_h$ increases, but the effect of $E$ on cost varies
across horizon ratios. For example, increasing $E$ from 5 to 10 reduces cost
by 34.43\% at $K_h=4$ and increases it by 12.87\% at $K_h=10$. These
contrasting effects motivate joint selection of $H$ and $E$. The horizon
term of the surrogate model fits the measured costs with a mean absolute relative error of 12.77\%
(Fig.~\ref{fig:compute-allocation}(d)), supporting our model of horizon error.

\subsection{The Cost Model Estimates a Test-Time Compute Strategy (Q3)}
\label{sec:compute-optimal-validation}

We now use the surrogate cost model to derive an approximate compute-optimal
strategy and ask whether it remains effective
across BFMs and motions.
To fit the full model, we combine configuration--cost pairs from the sweeps
in Sec.~\ref{sec:compute-configuration-effects} with measurements from
additional jointly sampled configurations to capture interactions between
solver and time horizon parameters, yielding 183 configuration--cost pairs in total.
For comparison, we also fit Random Forest, Radial Basis Function (RBF), and Gaussian Process
baselines on these data. At each budget $K$, we evaluate the
configuration proposed by each of the four models alongside multiple random configurations
satisfying $0.9K\leq K(\xi)\leq K$. We compare strategies using Regret: the percentage by which
a strategy's cost exceeds that of the best measured configuration (Empirical Optimum) at the same budget.
Appendix~\ref{app:compute-strategy-evaluation}
describes the fitting and evaluation procedures.

\noindent\textbf{The surrogate cost model identifies a near-optimal strategy in the fitting regime.}\
We compare the proposed configuration with 200 random configurations at the same
compute budget, on SONIC and on the motions used to fit the cost model.
Our strategy achieves the lowest mean regret of 4.02\% among the four strategies across six budgets, compared with 18.44\% for random forest, the next-best strategy.
Scaling test-time compute from $K=256$ to $K=8192$ consistently lowers the measured cost (Fig.~\ref{fig:compute-strategy}) under the proposed strategy, a total reduction of 58.0\%.

\begin{figure}[!tb]
    \centering
    \setlength{\abovecaptionskip}{4pt}
    \includegraphics[width=\linewidth]{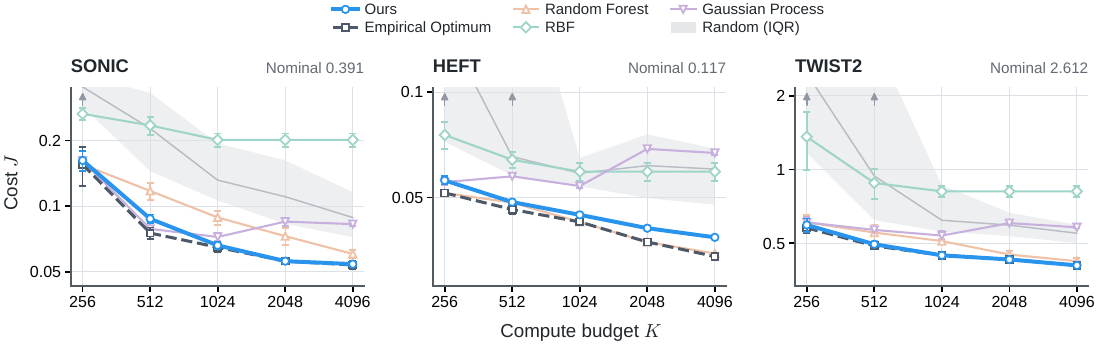}
    \caption{\textbf{The proposed strategy generalizes to unseen motions and BFMs.} Our strategy stays close to the empirical optimum and achieves the lowest average regret among the four strategies across three BFMs and five compute budgets. Faint lines and
    bands show median costs and 25th--75th percentiles across random configurations; nominal-command costs are labeled above the
    panels. Costs are averaged over five motions and three seeds;
    error bars show standard deviations of seed means. Each panel uses logarithmic axes.}
    \label{fig:compute-generalization}
\end{figure}

\noindent\textbf{The proposed strategy generalizes across BFMs and motions.}\
We further apply the proposed strategy unchanged to SONIC, HEFT, and TWIST2 on five new motions, comparing its proposed configurations with 50 random configurations (Fig.~\ref{fig:compute-generalization}).
Across the three BFMs and five budgets, our strategy achieves the lowest mean regret: 4.60\%, compared with 12.27\% for random forest. Its mean regrets are 4.72\%, 8.24\%, and 0.86\% on SONIC, HEFT, and TWIST2, respectively, making it the only strategy below 10\% on all three BFMs.
These results indicate that our
surrogate cost model, fitted to limited measurements from a single BFM, can select near-optimal configurations outside its fitting regime.

\subsection{The Benefits of Test-Time Compute Extend to Loco-Manipulation (Q4)}
\label{sec:loco-manipulation}

\begingroup
\setlength{\intextsep}{0pt}
\begin{wrapfigure}{R}{0.38\linewidth}
    \centering
    \setlength{\abovecaptionskip}{4pt}
    \includegraphics[width=0.92\linewidth]{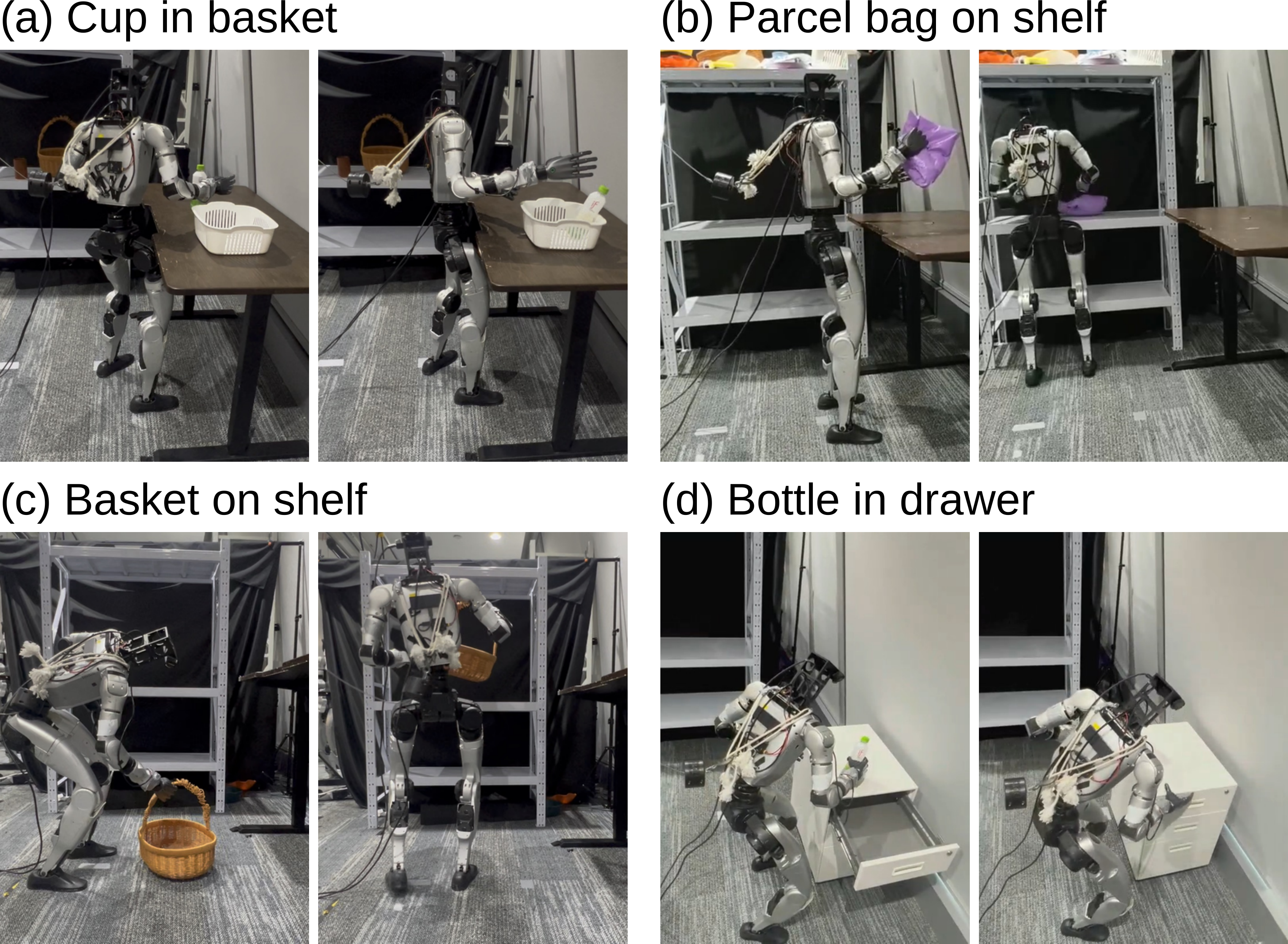}
    \caption{\raggedright\textbf{Illustration of real-robot loco-manipulation tasks.}}
    \label{fig:downstream-real-tasks}
\end{wrapfigure}

A high-level humanoid policy maps visual observations and optional language
instructions to motion commands executed by a BFM. Learning this mapping
requires demonstration labels compatible with that interface
(Appendix~\ref{app:high-level-policy}). We test whether optimized commands can serve as effective labels in both simulation and real-robot scenarios.
We keep demonstration images, proprioception, and hand labels fixed and
compare three body-action label sources:
\emph{original commands} issued during collection (Original), recorded \emph{robot states}
expressed as commands (State), and \emph{optimized commands} (Ours). For Ours,
we optimize commands offline with the BFM that will execute the learned policy's
commands (target BFM), using the recorded robot motion as the reference.
We randomize simulation dynamics during optimization to improve robustness
(Appendix~\ref{app:downstream-randomization}).
We also test cross-BFM settings, where the BFM used to collect demonstrations
(source BFM) differs from the target BFM.

For each label source, we train Diffusion Policy~\citep{chi2023diffusionpolicy}
on five HumanoidArena tasks~\citep{wang2026humanoidarena} following the benchmark protocol
(Appendix~\ref{app:downstream-simulation}), and fine-tune GR00T N1.7~\citep{nvidia2026gr00tn17}
on SONIC demonstrations of four real-robot loco-manipulation tasks
(Fig.~\ref{fig:downstream-real-tasks}; Appendix~\ref{app:downstream-hardware}).

\begin{table}[!tbp]
    \centering
    \setlength{\abovecaptionskip}{4pt}
    \renewcommand{\arraystretch}{1.05}
    \caption{\textbf{Simulation task success rates (\%) on HumanoidArena.}
    Each arrow points from the source BFM to the target BFM.
    Results are reported as mean $\pm$ sample standard deviation across three
    training seeds, with 60 evaluation rollouts per task and seed.}
    \label{tab:downstream-matched}
    \fontsize{9}{10}\selectfont
    \setlength{\tabcolsep}{4pt}
    \begingroup
\setlength{\aboverulesep}{0.7pt}
\setlength{\belowrulesep}{0.7pt}
\begin{tabular}{lrrrrr>{\hspace{1pt}}r}
\toprule
Labels & Boxing & Open Door & Pick \& Place Box & Sit Sofa & Vision Navigation & Mean \\
\midrule
\multicolumn{7}{@{}l}{\textbf{SONIC $\rightarrow$ SONIC}} \\
Original & $83.9_{\pm 1.0}$ & $84.4_{\pm 1.9}$ & $61.1_{\pm 1.0}$ & $88.9_{\pm 1.0}$ & $60.6_{\pm 1.0}$ & $75.8_{\pm 0.5}$ \\
\rowcolor[gray]{0.94}
\textbf{Ours} & $63.9_{\pm 1.0}$ & $51.1_{\pm 1.0}$ & $54.4_{\pm 1.0}$ & $66.7_{\pm 1.7}$ & $47.2_{\pm 1.0}$ & $56.7_{\pm 0.3}$ \\
State & $0.0_{\pm 0.0}$ & $0.0_{\pm 0.0}$ & $0.0_{\pm 0.0}$ & $0.0_{\pm 0.0}$ & $0.0_{\pm 0.0}$ & $0.0_{\pm 0.0}$ \\
\midrule
\multicolumn{7}{@{}l}{\textbf{SONIC $\rightarrow$ TWIST2}} \\
Original & $48.9_{\pm 1.9}$ & $0.0_{\pm 0.0}$ & $0.0_{\pm 0.0}$ & $17.8_{\pm 1.9}$ & $11.7_{\pm 1.7}$ & $15.7_{\pm 0.6}$ \\
\rowcolor[gray]{0.94}
\textbf{Ours} & $50.0_{\pm 1.7}$ & $0.0_{\pm 0.0}$ & $0.0_{\pm 0.0}$ & $21.7_{\pm 1.7}$ & $7.2_{\pm 1.9}$ & $15.8_{\pm 0.4}$ \\
State & $33.9_{\pm 1.0}$ & $0.0_{\pm 0.0}$ & $0.0_{\pm 0.0}$ & $14.4_{\pm 1.0}$ & $1.7_{\pm 1.7}$ & $10.0_{\pm 0.3}$ \\
\midrule
\multicolumn{7}{@{}l}{\textbf{SONIC $\rightarrow$ HEFT}} \\
Original & $46.7_{\pm 1.7}$ & $0.0_{\pm 0.0}$ & $0.0_{\pm 0.0}$ & $22.8_{\pm 1.9}$ & $16.1_{\pm 1.0}$ & $17.1_{\pm 0.4}$ \\
\rowcolor[gray]{0.94}
\textbf{Ours} & $57.2_{\pm 1.0}$ & $0.0_{\pm 0.0}$ & $0.0_{\pm 0.0}$ & $67.8_{\pm 1.0}$ & $18.9_{\pm 1.0}$ & $28.8_{\pm 0.2}$ \\
State & $38.9_{\pm 1.9}$ & $0.0_{\pm 0.0}$ & $0.0_{\pm 0.0}$ & $36.1_{\pm 1.0}$ & $27.8_{\pm 1.9}$ & $20.6_{\pm 0.7}$ \\
\bottomrule
\end{tabular}
\endgroup

    \setlength{\tabcolsep}{3.5pt}
    \par
    \normalsize
    \caption{\textbf{Real-robot task success rates (\%).}
    Each arrow points from the source BFM to the target BFM.
    Each task entry reports the mean success rate over 60 evaluation rollouts.}
    \label{tab:downstream-transfer}
    \fontsize{9}{10}\selectfont
    \begingroup
\setlength{\aboverulesep}{0.7pt}
\setlength{\belowrulesep}{0.7pt}
\begin{tabular}{@{}lr>{\columncolor[gray]{0.94}}rr@{\hspace{10pt}}r>{\columncolor[gray]{0.94}}rr@{}}
\toprule
& \multicolumn{3}{c}{\textbf{SONIC $\rightarrow$ SONIC}}
& \multicolumn{3}{c}{\textbf{SONIC $\rightarrow$ HEFT}} \\
\cmidrule(lr){2-4}\cmidrule(l){5-7}
Task & Original & \textbf{Ours} & State & Original & \textbf{Ours} & State \\
\midrule
Cup in basket & 55.0 & 55.0 & 0.0 & 0.0 & 40.0 & 13.3 \\
Parcel bag on shelf & 83.3 & 91.7 & 10.0 & 6.7 & 55.0 & 20.0 \\
Basket on shelf & 53.3 & 50.0 & 0.0 & 0.0 & 50.0 & 8.3 \\
Bottle in drawer & 50.0 & 60.0 & 0.0 & 0.0 & 50.0 & 6.7 \\
\midrule
Mean & 60.4 & 64.2 & 2.5 & 1.7 & 48.8 & 12.1 \\
\bottomrule
\end{tabular}
\endgroup

\end{table}

\noindent\textbf{Optimized commands serve as effective action labels.}\
When the source and target BFMs are the same, policies trained on original
commands provide a performance baseline because those commands generated the
recorded motion. Command optimization uses the frozen BFM to infer alternative
command labels from that motion.
Policies trained on these labels achieve higher mean success rates than
those trained on robot-state labels and approach the performance of policies
trained on original commands in simulation and on hardware
(Tables~\ref{tab:downstream-matched} and~\ref{tab:downstream-transfer}).
This performance recovery supports optimized commands as a partial
substitute for the original action labels.

\noindent\textbf{Command optimization enables demonstration reuse across BFMs.}\
When the target BFM changes, re-optimizing the command labels preserves
the recorded motion as the reference while adapting supervision to the new
controller. The resulting policies outperform or approach both
original-command and robot-state baselines in mean success rate for most simulation pairings and  on all four real-robot tasks
(Tables~\ref{tab:downstream-transfer} and ~\ref{tab:downstream-full}). They also retain part of the success rate
of policies trained on optimized labels for the source BFM and evaluated with that same BFM.
This indicates that label adaptation reduces dependence on the source BFM.
\par\endgroup

\section{Conclusion}
\label{sec:conclusion}

Our study shows that an existing humanoid BFM can track a motion more
accurately when its command differs from the desired motion. Test-Time
Command Optimization finds such commands, with gains governed by both
compute budget and allocation. The same optimization recovers useful action
labels from recorded motions, extending the benefits to downstream learning
and demonstration reuse across BFMs. By bringing the distinction between
desired motion and input command into focus, our study reveals substantial
room to improve existing BFMs through test-time compute.

\noindent\textbf{Limitations.}\ Our current implementation is not yet fast
enough to deliver satisfactory real-time performance, limiting our study of
closed-loop command optimization on real robots. Optimizing motion commands
improves tracking accuracy but can introduce temporal irregularities in the
commands and jitter in the executed motion. Residual execution errors can
still limit downstream tasks that require precise contact and coordinated
whole-body movement.

\subsection*{AI use statement}

In this work, we used generative AI tools for designing or providing feedback on
research methodology or experiments, implementing methods, assisting with
translation, cleaning and reformatting datasets, and supporting qualitative and
thematic data analysis. We have not used generative AI tools for generating
synthetic datasets, helping develop theoretical models or conceptual frameworks,
formulating mathematical claims, providing critical ingredients for proving
mathematical claims, assisting in the writing of proofs, proposing or refining
hypotheses, or interpreting results. Additionally, we used generative AI tools
for creating or modifying scientific figures or images, creating or editing software code, drafting parts
of the paper, and editing the paper to improve readability. We have reviewed all
AI-assisted work. We take responsibility for the final content of this work,
including text, claims or artifacts produced with the aid of generative AI.

\subsection*{Reproducibility statement}

To ensure the reproducibility of our work, we provide the code and example data as part of the supplementary material.
We further expand upon the experimental details explained in the main text in Appendix~\ref{app:method-details} and~\ref{app:experimental-setup-and-results}.

\bibliographystyle{plainnat}
\bibliography{references}

\appendix
\clearpage
\raggedbottom
\addtocontents{toc}{\protect\setcounter{tocdepth}{3}}
\begingroup
\renewcommand{\contentsname}{Appendix Contents}
\small
\setlength{\parskip}{0pt}
\tableofcontents
\endgroup
\clearpage
\section{Preliminaries}
\label{app:preliminaries}

Hierarchical humanoid control connects two learned policies: a motion
controller converts motion commands and robot feedback into control actions,
while a visuomotor policy predicts those commands from sensory observations
and task instructions~\citep{ze2025twist2,luo2026sonic,hu2026openhlm}. The following formulation describes how each policy
is learned and how their interfaces connect.

\subsection{Humanoid motion tracking}
\label{app:general-motion-tracking}

Humanoid motion tracking is the task of reproducing a reference motion
through feedback control under the robot's dynamics. A reference
$\tau^{\mathrm{ref}}_{1:T}$ specifies the desired evolution of the robot's
configuration and velocity. In the representation of
Section~\ref{sec:method-preliminaries}, each frame has $2n_q+13$ entries:
three root-position coordinates, a unit quaternion for root orientation,
$n_q$ joint positions, and the corresponding three-dimensional root linear
and angular velocities and $n_q$ joint velocities. Human motion data can
supply these references after retargeting to the robot's kinematics;
body positions and orientations follow from forward kinematics
\citep{liao2025beyondmimic,luo2026sonic}.
The actual system state $s_t$ evolves during execution. The policy receives
feedback observations $o_t$, such as joint positions, joint velocities,
and inertial measurements, rather than direct access to every state variable~\citep{liao2025beyondmimic,ze2025twist2}.

A motion command $c$ supplies the movement to execute. Combined with feedback,
it determines the low-level action through
\begin{equation}
    a_t=\pi(o_{0:t-1},c_t),\qquad
    (s_t,o_t)=f(s_{t-1},a_t).
    \label{eq:appendix-tracking-loop}
\end{equation}
Here $t$ indexes the target frame reached from state $s_{t-1}$.
The policy interface converts physical motion coordinates into its native
inputs through coordinate transforms, feature construction, or learned
encoding~\citep{liao2025beyondmimic,luo2026sonic}. The compact notation suppresses controller-specific extraction of
temporal context from $c$: an input may encode a motion window rather than
only the current frame~\citep{luo2026sonic}.
For controllers with joint-position actions, $a_t$ specifies setpoints that
proportional--derivative joint controllers convert into actuator torques
\citep{liao2025beyondmimic}. This actuation step is included in $f$.
Thus, $c_t$ specifies the movement to reproduce, whereas $a_t$ is the control
output selected from the current feedback to realize it. Repeated feedback
updates produce the executed trajectory $\Tau_\pi(c)$.

Motion-tracking controllers are commonly learned through reinforcement
learning in physics simulation~\citep{liao2025beyondmimic,luo2026sonic,ze2025twist2}. Training samples reference motions and
starting phases, initializes the robot, and updates the policy from its
simulated execution~\citep{liao2025beyondmimic}. The sampled reference supplies both the motion input
and the tracking target. With $p_{\mathrm{train}}$ denoting the training-motion
distribution, a standard objective is
\begin{equation}
    \max_{\pi}\;
    \mathbb E_{\tau^{\mathrm{ref}}\sim p_{\mathrm{train}},\,\pi,\,f}
    \left[\sum_{t=1}^{T}\gamma^{t-1}
    \bigl(r_t^{\mathrm{track}}+r_t^{\mathrm{reg}}\bigr)\right],
    \label{eq:tracking-pretraining}
\end{equation}
where $\gamma\in(0,1]$ is a discount factor and the expectation also includes sampled
initial conditions and training perturbations. Tracking rewards compare
simulated and reference positions, orientations, or velocities; regularization
terms can penalize abrupt action changes, joint-limit violations, and
undesired contacts~\citep{liao2025beyondmimic,luo2026sonic}.
Local pose matching and global position tracking constrain different aspects
of the motion, so the tracked quantities and coordinate frames are part of
the training objective~\citep{zeng2026scalebfm,zhu2026clot}.
Dynamics randomization and observation perturbations expose the controller
to variability encountered during deployment~\citep{liao2025beyondmimic,ze2025twist2}. This objective trains the
controller's parameters; $J$ in Section~\ref{sec:method-preliminaries}
evaluates commands for an already frozen controller.

Training a shared controller across diverse motions makes whole-body tracking
a reusable capability. Humanoid Behavior Foundation Models (BFMs) develop
this capability through broad motion training and expose control interfaces
that support different motion specifications
\citep{zeng2025bfm,zeng2026scalebfm,luo2026sonic}.
The pretrained policy supplies the low-level control needed to execute new
motion inputs, allowing higher-level policies to learn task-dependent
commands while keeping the motion controller fixed~\citep{ze2025twist2,luo2026sonic}.

\subsection{Learning Humanoid Visuomotor Policies}
\label{app:high-level-policy}

A humanoid visuomotor policy learns to select motion commands from visual
input and robot-related observations~\citep{ze2025twist2}. A vision-language-action (VLA) model
additionally conditions on a language instruction specifying the task
\citep{bjorck2025gr00tn1}. In a hierarchical system, these predictions drive
the motion controller through its command interface
\citep{ze2025twist2,luo2026sonic}.
Let $y_t$ denote high-level observations, $\ell$ an optional instruction,
and $u_t$ the high-level action. The body component of $u_t$ represents a
motion command or its encoding; any hand actions are sent through the
corresponding hand-control interface. The BFM's output $a_t$ remains the
low-level control action.

Demonstrations provide paired observations and action labels for learning
this mapping. During teleoperation, the sensory observations and commands
issued to the robot can be recorded together~\citep{ze2025twist2}.
The issued command and the motion actually achieved are distinct records:
one specifies the controller's input, while the other describes its physical
response. Consequently, an action label must be interpreted in the input
representation of the controller that will execute it~\citep{ze2025twist2,hu2026openhlm}. This representation
may contain physical motion targets or learned motion tokens
\citep{ze2025twist2,luo2026sonic}.

Supervised policy learning fits the relationship between demonstration
observations and action sequences~\citep{chi2023diffusionpolicy}. For a prediction length $L$ on the
high-level time grid, a conditional policy generates an action chunk as
\begin{equation}
    u_{t:t+L-1}\sim p_\theta(\,\cdot\mid y_{1:t},\ell).
    \label{eq:high-level-policy}
\end{equation}
Here $\theta$ denotes the high-level policy parameters learned from the
demonstrations. Diffusion policies learn conditional action sequences through
denoising~\citep{chi2023diffusionpolicy}; VLA models can combine visual and
language representations with generative action prediction
\citep{bjorck2025gr00tn1}.

At inference, a prefix of the predicted chunk is executed before the policy
is queried with updated observations. Body actions are converted to the BFM's
input representation and control-time grid, while the BFM continually updates
its output from robot feedback~\citep{chi2023diffusionpolicy,ze2025twist2}.
The high-level prediction length $L$ and execution interval describe this
policy's inference schedule; they are separate from the command optimizer's
horizon $H$ and replanning interval $E$. Concrete action representations and
deployment protocols are given in Appendix~\ref{app:downstream}.

\clearpage
\section{Method Details}
\label{app:method-details}

This appendix specifies the rollout cost, command parameterization,
controller inputs, and sampling optimizers, then describes compute allocation
under the fitted surrogate cost model.

\subsection{Rollout Cost}
\label{app:rollout-cost}

For a robot with $n_q$ actuated joints, let $q^A$ and $\dot q^A$ denote
the joint positions and velocities in motion frame $A$, and use the same
notation for frame $B$. For each of the $B_0$ evaluated bodies, let $p_b$
be its world position and $u_b$ its unit quaternion. The joint and body
discrepancies are
\begin{align}
    e_q(A,B)&=\frac{1}{n_q}\sum_{j=1}^{n_q}(q_j^A-q_j^B)^2,
    &e_{\dot q}(A,B)&=\frac{1}{n_q}\sum_{j=1}^{n_q}(\dot q_j^A-\dot q_j^B)^2,\\
    e_x(A,B)&=\frac{1}{B_0}\sum_{b=1}^{B_0}\lVert p_b^A-p_b^B\rVert_2^2,
    &e_R(A,B)&=\frac{1}{B_0}\sum_{b=1}^{B_0}
      \left[2\arccos\!\left(\min(1,|(u_b^A)^\top u_b^B|)\right)\right]^2.
    \label{eq:app-frame-errors}
\end{align}
Body-position error sums the three squared coordinate errors before
averaging over bodies. The absolute quaternion inner product makes
orientation error invariant to quaternion sign.

Tracking errors compare the executed robot frame with the reference frame;
deformation errors compare the decoded command frame with the reference.
Body positions and orientations come from forward kinematics and are
expressed in world coordinates. Using superscripts $\mathrm{track}$ and $\mathrm{deform}$
for these comparisons, the running and terminal terms in
Eq.~\ref{eq:finite-horizon-cost} are
\begin{align}
    \ell_t^{\mathrm{track}}
    &=w_q e_{q,t}^{\mathrm{track}}
      +w_{\dot q}e_{\dot q,t}^{\mathrm{track}}
      +w_xe_{x,t}^{\mathrm{track}}
      +w_Re_{R,t}^{\mathrm{track}},\\
    \ell_t^{\mathrm{reg}}
    &=\bar w_q e_{q,t}^{\mathrm{deform}}
      +\bar w_x e_{x,t}^{\mathrm{deform}}
      +\bar w_R e_{R,t}^{\mathrm{deform}}
      +\frac{w_s}{n_q}\lVert\dot q_t\rVert_2^2,\\
    \ell_{t+H-1}^{\mathrm{term}}
    &=w_q^{\mathrm{term}} e_{q,t+H-1}^{\mathrm{track}}
      +w_x^{\mathrm{term}} e_{x,t+H-1}^{\mathrm{track}}.
\end{align}
Here $\dot q_t$ denotes the executed robot's joint velocities.
Each running loss $\ell_t$ scores the state reached after applying $c_t$
for one control period against reference frame $\tau_t^{\mathrm{ref}}$.
The terminal loss uses the same state and reference frame as the last
running loss, at $t+H-1$. The full-sequence cost $J$ sums the running losses
over executed steps.

We use the following cost weights:
\begin{equation}
    \begin{aligned}
        (w_q,w_{\dot q},w_x,w_R)&=(1,0,1,0.5),
        & (\bar w_q,\bar w_x,\bar w_R)&=(0.1,0.1,0.05),\\
        w_s&=0.01,
        & (w_q^{\mathrm{term}},w_x^{\mathrm{term}})&=(5,5).
    \end{aligned}
    \label{eq:app-tracking-cost-weights}
\end{equation}

\subsection{Command Parameterization and Controller Inputs}
\label{app:command-implementation}

We optimize offsets from the nominal command using
$n_{\mathrm{knot}}$ control knots. For an $H$-step plan, we set
\begin{equation}
    n_{\mathrm{knot}}(H)
    =\max\!\left(2,\left\lfloor\frac{5H}{40}+\frac12\right\rfloor\right).
    \label{eq:app-command-knots}
\end{equation}
Each row of the decision array $x$
represents one knot, with root-position offsets, a three-dimensional
rotation vector, and joint-position offsets. The knots are uniformly
spaced at $\kappa_j=j(H-1)/(n_{\mathrm{knot}}-1)$,
$j=0,\ldots,n_{\mathrm{knot}}-1$.

We clip the knot coordinates to $[-1,1]$, multiply them by the diagonal
matrix of physical scales $D$, and linearly interpolate them to the $H$
control steps with $L_H$. The resulting command offsets are
\begin{equation}
    (\Delta p_h,\Delta r_h,\Delta q_h)
    =\left[L_H\!\left(\operatorname{clip}(x,-1,1)D\right)\right]_h,
    \qquad h=0,\ldots,H-1.
\end{equation}
Clipping applies when constructing commands; optimizer updates use the
original knot coordinates and the costs of their decoded commands.
At each step, we add $\Delta p_h$ and $\Delta q_h$ to that step's nominal
root and joint positions.
The rotation offset is converted to a quaternion and left-multiplied with
the nominal root orientation,
$u_h^c=\operatorname{quat}(\Delta r_h)\otimes u_h^{\mathrm{ref}}$.
Here $\operatorname{quat}$ maps a rotation vector to a unit quaternion,
and $\otimes$ denotes quaternion multiplication. Forward kinematics then
gives the body poses.

With control period $\Delta t$, we derive command velocities from the
decoded poses. For a joint coordinate or root-position coordinate $v_h$,
\begin{equation}
    \dot v_h=\frac{v_{r_h}-v_{l_h}}{(r_h-l_h)\Delta t},
    \qquad l_h=\max(h-1,0),\quad r_h=\min(h+1,H-1).
\end{equation}
This uses centered differences in the interior and one-sided differences
at the endpoints. Root angular velocity comes from the quaternion difference
over the same interval; forward kinematics provides body linear and angular
velocities.

\subsection{Sampling Optimizers}
\label{app:sampling-optimizers}

The first planning call starts with mean $\mu=0$. Each later call initializes
the mean by shifting the previous best candidate's clipped knot offsets
forward by $E$ steps, using linear interpolation and a zero-filled tail.
The remaining optimizer state is reset at each planning call.

\noindent\textbf{CEM~\citep{deboer2005cem}.}\ At each iteration, we draw
candidate knot arrays $x_n=\mu+\sigma\odot\epsilon_n$ from a Gaussian with
coordinate-wise mean $\mu$ and standard deviation $\sigma$.
The noise arrays $\epsilon_n\sim\mathcal N(0,\mathbf I)$ are independent,
and $\odot$ denotes element-wise multiplication. The $M$ lowest-cost
candidates form the elite set $\mathcal E$, which updates the distribution:
\begin{equation}
    \mu^+=\frac{1}{M}\sum_{n\in\mathcal E}x_n,
    \qquad
    \sigma^+=\max\!\left(\sigma_{\min},
      \sqrt{\frac{1}{M}\sum_{n\in\mathcal E}(x_n-\mu^+)^2}\right).
\end{equation}
Here $\sigma_{\min}$ is the minimum standard deviation. All operations
act coordinate-wise, and the variance uses divisor $M$.

\noindent\textbf{Predictive sampling~\citep{howell2022predictive}.}\ We draw
candidate knot arrays $x_n=\mu+\sigma_{\mathrm{PS}}\epsilon_n$, with fixed
noise scale $\sigma_{\mathrm{PS}}$ and independent standard Gaussian arrays
$\epsilon_n$. The current mean is included as $x_1=\mu$.
If $J_n$ is a candidate's sum of running and terminal costs, the
lowest-cost candidate becomes the next mean:
$\mu^+=x_{n^*}$, where $n^*=\operatorname*{argmin}_n J_n$.
The noise scale stays fixed within and across planning calls.

\noindent\textbf{MPPI-CMA~\citep{jordana2025zeroorder}.}\ Each knot $j$ has
a mean $\mu_j$ and a full covariance $\Sigma_j$. Knots are sampled
independently, $x_{n,j}\sim\mathcal N(\mu_j,\Sigma_j)$, so the complete
covariance is block diagonal across knots. Each candidate's summed running
and terminal cost $J_n$ determines its weight,
\begin{equation}
    w_n=\frac{\exp[-(J_n-J_{\min})/\vartheta]}
    {\sum_{m=1}^{N}\exp[-(J_m-J_{\min})/\vartheta]},
    \qquad J_{\min}=\min_n J_n.
    \label{eq:app-sampling-weights}
\end{equation}
where temperature $\vartheta$ controls how strongly the weights favor
lower costs. With adaptation rate $\rho$, the mean and covariance become
\begin{equation}
    \begin{aligned}
        \mu_j^+&=\sum_{n=1}^{N}w_nx_{n,j},\\
        \widetilde\Sigma_j
        &=(1-\rho)\Sigma_j+
          \rho\sum_{n=1}^{N}w_n(x_{n,j}-\mu_j)(x_{n,j}-\mu_j)^\top.
    \end{aligned}
\end{equation}
The covariance update uses deviations from the previous mean $\mu_j$.
The next covariance $\Sigma_j^+$ floors each eigenvalue of
$\widetilde\Sigma_j$ at $\sigma_{\min}^2$, where $\sigma_{\min}$ is the
minimum noise scale. Each planning call starts from an isotropic covariance.

\noindent\textbf{CMA-ES~\citep{hansen2001cmaes}.}\ The separable
variant~\citep{ros2008sepcmaes} optimizes all control-knot coordinates together.
Its population contains $N\ge2$ candidates. Sampling and updates use the unclipped
population and its costs, retaining the evolution state within each
planning call.

\noindent\textbf{CBO~\citep{sun2026cbo}.}\ The initial particles are
$x_n=\mu+\sigma_{\mathrm{init}}\epsilon_n$, with noise scale
$\sigma_{\mathrm{init}}$ and independent standard Gaussian arrays
$\epsilon_n$. At each iteration, a particle's summed running and terminal
cost $J_n$ gives weight $w_n\propto\exp[-(J_n-J_{\min})/\vartheta]$,
normalized so that $\sum_n w_n=1$. Here $J_{\min}=\min_n J_n$ and
$\vartheta$ is the temperature. The weighted mean
$\bar x=\sum_n w_nx_n$ is the consensus. With independent standard
Gaussian arrays $\zeta_n$, the next particles are
\begin{equation}
    x_n^+=x_n-\lambda\Delta s(x_n-\bar x)
      +\sigma_{\mathrm{CBO}}\sqrt{\Delta s}\,
       |x_n-\bar x|\odot\zeta_n,
\end{equation}
Here $\lambda$ is the consensus weight, $\sigma_{\mathrm{CBO}}$ is the
noise weight, and $\Delta s$ is the step size.
The absolute value acts coordinate-wise. The updated particles are
evaluated in the next iteration; the consensus is stored as the
optimizer mean. Execution uses the best evaluated particle across
iterations.

\subsection{Allocation Under the Fitted Model}
\label{app:compute-optimal-strategy}

Let $J_{\mathrm{TTC}}(\mathcal A,\xi,\tau)$ denote the expected full-sequence
cost on a motion $\tau$ of length $T_\tau$. Its time-normalized mean over
the reference-motion dataset $\mathcal D$ is
\begin{equation}
    J_{\mathrm{TTC}}(\mathcal A,\xi)
    =\frac{1}{|\mathcal D|}
    \sum_{\tau\in\mathcal D}
    J_{\mathrm{TTC}}(\mathcal A,\xi,\tau)/T_\tau.
    \label{eq:dataset-ttc-objective}
\end{equation}
The expectation is over the sampling optimizer. Equation~\ref{eq:compute-optimal-configuration}
defines the algorithm and allocation that minimize this dataset cost
under each compute budget.

For a fixed sampling optimizer and its fitted surrogate parameters,
consider a solver budget $O=IN$ at fixed $H,E$. The solver term becomes
\begin{equation}
    P_O(I)=C_I I^{-\alpha_I}+C_N O^{-\alpha_N}I^{\alpha_N}.
\end{equation}
The horizon cost is constant, and the sampling term is a strictly
increasing function of $P_O$. In a continuous relaxation with $I>0$,
setting the derivative to zero gives
\begin{equation}
    I_{\mathrm{cont}}^*
      =\left(\frac{\alpha_I C_I}{\alpha_N C_N}\right)^{1/(\alpha_I+\alpha_N)}
       O^{\alpha_N/(\alpha_I+\alpha_N)},
    \qquad N_{\mathrm{cont}}^*=\frac{O}{I_{\mathrm{cont}}^*}.
    \label{eq:app-continuous-solver-split}
\end{equation}
As a function of $\log I$, $P_O$ is the sum of two strictly convex
exponentials, so this stationary point is its unique minimum.

For integer configurations, define
$P=C_I I^{-\alpha_I}+C_N N^{-\alpha_N}>0$. At fixed $(I,H,E)$,
\begin{equation}
    \frac{\partial\widehat J}{\partial N}
      =-\frac{\eta\alpha_N C_N}{H}\gamma^H
         P^{\eta/H-1}N^{-\alpha_N-1}<0.
\end{equation}
All fitted coefficients and exponents are positive. We therefore
enumerate feasible integer triples $(I,H,E)$ and evaluate the
largest candidate count allowed by the budget and resource constraints.
The configuration with the lowest predicted cost among these triples
is the allocation selected by the fitted model.

\clearpage
\section{Additional Experimental Setup and Results}
\label{app:experimental-setup-and-results}

We begin with the shared simulation setup
(Appendix~\ref{app:tracking-simulation-setup}), followed by motion-tracking
results (Appendix~\ref{app:large-scale-motion-tracking}) and the effects of
solver and time horizon parameters (Appendix~\ref{app:compute-allocation-effects}).
The cost-model experiments in Appendix~\ref{app:compute-strategy-evaluation}
cover fitting, configuration selection, evaluation metrics, and tests on
fitting motions and new motions across BFMs.
Appendix~\ref{app:downstream} gives the loco-manipulation data, training,
and evaluation protocols for simulation and hardware, followed by the
complete simulation results.

\subsection{Experimental Setup}
\label{app:tracking-simulation-setup}

\noindent\textbf{Robot and simulation.}\ The G1 has 29 actuated joints.
The physics time step is $0.005$\,s, and each BFM runs at 50\,Hz with four
physics steps per action. Thus the control period is $\Delta t=0.02$\,s.

We use the official released SONIC G1~\citep{luo2026sonic}, HEFT PMG~\citep{liu2026heft},
and TWIST2 1017--20k~\citep{ze2025twist2} checkpoints.

\noindent\textbf{Command settings.}\ Each control knot has 35 coordinates.
The physical scales in Appendix~\ref{app:command-implementation}
are $0.055$\,m for each horizontal root coordinate,
$0.0275$\,m for root height, $0.0825$\,rad for each rotation-vector
coordinate, and $0.1375$\,rad for each joint.

\subsection{Test-Time Command Optimization Improves Tracking Precision (Q1)}
\label{app:large-scale-motion-tracking}

\subsubsection{AMASS Evaluation and Tracking Metrics}
\label{app:amass-protocol}

\noindent\textbf{Evaluation protocol.}\ Each BFM executes all 12,273 retargeted
AMASS motions once with nominal commands and once with optimized commands.
Each pair starts from the first reference frame with identical
robot states, controller states, and observation histories.
We use CEM with $(I,N,H,E)=(4,80,40,20)$, nine elites, and the noise
scales in Table~\ref{tab:solver-settings}.
For the mean metrics, we first average within each motion, then across
motions with equal weight.

\noindent\textbf{Cost-win rate.}\ This is the percentage of motions for which
optimized commands yield a lower cost than nominal commands.

\noindent\textbf{Position errors.}\ Global body-position error is the mean
Euclidean distance between executed and reference positions in world
coordinates, averaged over control steps and 14 bodies. These are the pelvis,
torso, and the left and right hip-roll, knee, ankle-roll, shoulder-roll,
elbow, and wrist-yaw links. Root-position error uses the pelvis alone.

\noindent\textbf{Joint acceleration.}\ For motion $m$ with $F_m$ frames,
let $T_m=F_m-1$ be the number of executed steps. We compute joint acceleration
from consecutive executed joint velocities with $\Delta t=0.02$\,s,
then average its absolute value over time and $n_q=29$ joints:
\begin{equation}
    \ddot q_{m,t,j}
    =\frac{\dot q_{m,t+1,j}-\dot q_{m,t,j}}{\Delta t},
    \qquad
    A_m=\frac{1}{n_q(T_m-1)}
        \sum_{t=1}^{T_m-1}\sum_{j=1}^{n_q}|\ddot q_{m,t,j}|.
    \label{eq:app-joint-acceleration}
\end{equation}
We report acceleration in rad/s$^2$.

\clearpage
\subsubsection{Per-Motion Tracking Precision and Cost}
\label{app:amass-additional-results}

\begin{figure}[!h]
    \centering
    \includegraphics[width=\linewidth]{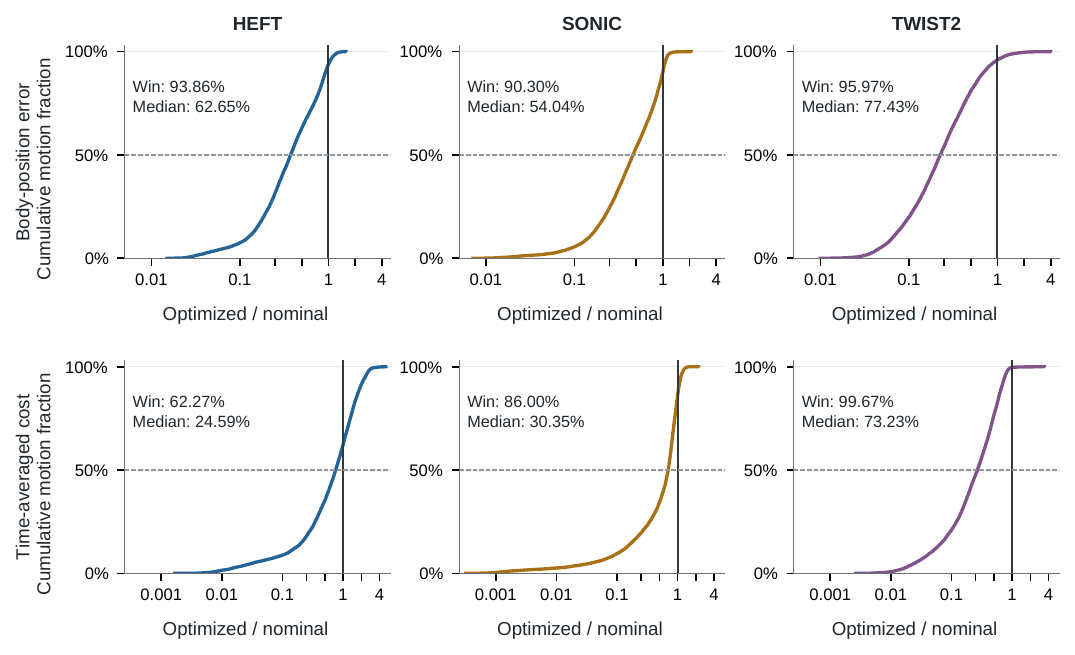}
    \caption{\textbf{Motion-level tracking performance on AMASS.}
    Empirical cumulative distributions of optimized-to-nominal global
    body-position error (top) and full-sequence cost per time step (bottom)
    across 12,273 motions per BFM. Ratios below one indicate improvement.
    Horizontal axes are logarithmic; reference lines mark a ratio of one and
    the 50th percentile. Panels report win rates and median percentage reductions.}
    \label{fig:amass-paired-ratios}
\end{figure}

Figure~\ref{fig:amass-paired-ratios} and Table~\ref{tab:amass-paired-distribution}
show the frequency and size of the improvements that contribute to the mean
reductions in Table~\ref{tab:exp-amass}.

\begin{table}[!h]
\centering
\footnotesize
\setlength{\tabcolsep}{3pt}
\caption{\textbf{Motion-level reductions and increases in tracking error and cost.}
Global body-position error and full-sequence cost per time step for 12,273
motions per BFM. Median reduction is the median of the per-motion percentage
reductions, $100(1-X_m^{\mathrm{opt}}/X_m^{\mathrm{nom}})$, for metric $X$.
Win rate is the percentage of motions with a lower error or cost after optimization.
$P_{95}$ is the 95th percentile of $X_m^{\mathrm{opt}}-X_m^{\mathrm{nom}}$
among motions with increased error or cost, in the metric's units.}
\label{tab:amass-paired-distribution}
\begin{tabular}{@{}llcccc@{}}
\toprule
BFM & Metric & \shortstack{Median\\nominal $\to$ optimized} &
\shortstack{Median\\reduction (\%)} & \shortstack{Win\\(\%)} &
\shortstack{Increase\\$P_{95}$} \\
\midrule
HEFT & Body-position error (cm) & $13.88\to 4.80$ & 62.65 & 93.86 & 1.20 \\
 & Cost & $0.0529\to 0.0367$ & 24.59 & 62.27 & 0.0148 \\
\addlinespace[2pt]
SONIC & Body-position error (cm) & $11.51\to 4.86$ & 54.04 & 90.30 & 1.09 \\
 & Cost & $0.0677\to 0.0416$ & 30.35 & 86.00 & 0.0054 \\
\addlinespace[2pt]
TWIST2 & Body-position error (cm) & $62.68\to 13.15$ & 77.43 & 95.97 & 23.76 \\
 & Cost & $1.8983\to 0.4709$ & 73.23 & 99.67 & 6.4616 \\
\bottomrule
\end{tabular}
\end{table}

To relate improvements to nominal tracking difficulty, we sort motions by
nominal global body-position error within each BFM and divide them into
four quartiles (Table~\ref{tab:amass-nominal-error-contributions}).
For each quartile, we report the median change and its contribution to
the total reductions and increases.

\begin{table}[!ht]
\centering
\footnotesize
\setlength{\tabcolsep}{3pt}
\caption{\textbf{Tracking performance grouped by nominal body-position error.}
Quartiles run from lowest (Q1) to highest (Q4) nominal global body-position
error within each BFM. Error reduction is nominal minus optimized error.
For each contribution column, we sum only changes in the indicated direction:
error decreases, error increases, or cost decreases.
We divide each group sum by the corresponding total over all motions of
the same BFM and express the result as a percentage.}
\label{tab:amass-nominal-error-contributions}
\begin{tabular}{@{}llrrrrr@{}}
\toprule
& & & & \multicolumn{3}{c}{Contribution to total (\%)} \\
\cmidrule(lr){5-7}
BFM & Group & \shortstack{Nominal error\\range (cm)} &
\shortstack{Median error\\reduction (cm)} & \shortstack{Error\\reduction} &
\shortstack{Error\\increase} & \shortstack{Cost\\reduction} \\
\midrule
HEFT & Q1 & 1.89--6.66 & 0.57 & 0.86 & 78.53 & 0.01 \\
 & Q2 & 6.66--13.88 & 5.11 & 6.00 & 21.33 & 0.26 \\
 & Q3 & 13.88--27.78 & 13.78 & 16.60 & 0.14 & 2.92 \\
 & Q4 & 27.79--843.14 & 38.04 & 76.53 & 0.00 & 96.81 \\
\addlinespace[3pt]
SONIC & Q1 & 1.89--5.63 & 0.22 & 0.23 & 43.60 & 0.02 \\
 & Q2 & 5.63--11.51 & 3.09 & 1.67 & 8.29 & 0.05 \\
 & Q3 & 11.51--28.56 & 11.64 & 6.33 & 3.73 & 0.23 \\
 & Q4 & 28.57--3065.46 & 60.81 & 91.77 & 44.38 & 99.70 \\
\addlinespace[3pt]
TWIST2 & Q1 & 3.72--31.43 & 7.87 & 2.27 & 39.10 & 2.29 \\
 & Q2 & 31.43--62.68 & 31.62 & 8.74 & 19.57 & 4.97 \\
 & Q3 & 62.71--130.24 & 70.86 & 19.74 & 26.58 & 11.28 \\
 & Q4 & 130.31--1412.99 & 206.07 & 69.25 & 14.75 & 81.45 \\
\bottomrule
\end{tabular}
\end{table}

\subsection{Compute Allocation Governs the Gains (Q2)}
\label{app:compute-allocation-effects}

Table~\ref{tab:allocation-motions} lists the ten motions used in
Sec.~\ref{sec:compute-configuration-effects}. Every configuration uses
three optimizer seeds. We estimate the dataset cost in
Eq.~\ref{eq:dataset-ttc-objective} by averaging the three seed means.

\begin{table}[t]
    \centering
    \caption{Reference motions used in the compute-allocation and strategy
    experiments.}
    \label{tab:allocation-motions}
    \footnotesize
    \setlength{\tabcolsep}{4pt}
    \begin{tabular}{@{}lrr@{}}
        \toprule
        Source sequence & Frames & Duration (s) \\
        \midrule
        \texttt{CMU/63/63\_01\_poses} & 200 & 3.98 \\
        \texttt{BMLmovi/Subject\_22\_F\_MoSh/Subject\_22\_F\_3\_poses} & 159 & 3.16 \\
        \texttt{KIT/883/wipe\_leg\_bigcircle03\_poses} & 267 & 5.32 \\
        \texttt{BMLmovi/Subject\_45\_F\_MoSh/Subject\_45\_F\_19\_poses} & 199 & 3.96 \\
        \texttt{BioMotionLab\_NTroje/rub061/0014\_knocking2\_poses} & 507 & 10.12 \\
        \texttt{CMU/102/102\_20\_poses} & 265 & 5.28 \\
        \texttt{KIT/8/WalkInClockwiseCircle01\_poses} & 464 & 9.26 \\
        \texttt{BioMotionLab\_NTroje/rub043/0010\_normal\_jog2\_poses} & 84 & 1.66 \\
        \texttt{KIT/424/bend\_left09\_poses} & 335 & 6.68 \\
        \texttt{CMU/62/62\_14\_poses} & 575 & 11.48 \\
        \bottomrule
    \end{tabular}
\end{table}

\subsubsection{Solver Parameters}
\label{app:solver-sweep}

We fix $(H,E)=(40,20)$ and evaluate solver budgets
$O\in\{256,512,1024,2048,4096,8192\}$. At each budget, we test
\begin{equation}
    I\in\{1,2,4,8,16,32,64,128,256,512\},\qquad
    I\le O,\qquad N=O/I.
    \label{eq:app-solver-grid}
\end{equation}
CMA-ES uses only configurations with $N\ge2$.
Table~\ref{tab:solver-settings} gives the optimizer settings.

\begin{table}[t]
    \centering
    \caption{Fixed optimizer settings in the solver sweep. Noise scales
    apply to the normalized knot coordinates before clipping and physical
    scaling. The CEM settings also apply to the horizon and strategy
    experiments.}
    \label{tab:solver-settings}
    \small
    \begin{tabular}{@{}lp{0.76\linewidth}@{}}
        \toprule
        Optimizer & Settings \\
        \midrule
        CEM & Initial standard deviation $1.6$; minimum standard deviation
        $0.075$; $\max(1,\lfloor0.1N\rfloor)$ elites; exploration fraction $0$. \\
        PS & Gaussian noise scale $1.2$. \\
        MPPI-CMA & Initial noise scale $1.6$; minimum noise scale $0.075$;
        temperature $0.1$; covariance adaptation rate $0.1$.
        Covariance eigenvalues are floored at $0.075^2$. \\
        CMA-ES & Separable CMA-ES; initial scale $1.6$; minimum scale
        $0.05$; maximum scale $2.0$. \\
        CBO & Initial noise scale $1.0$; temperature $0.1$; consensus
        weight $1.0$; noise weight $0.3$; step size $0.1$. \\
        \bottomrule
    \end{tabular}
\end{table}

\subsubsection{Time Horizon Parameters}
\label{app:horizon-sweep}

We evaluate all 24 pairs formed by
$E\in\{5,10,20,40\}$ and $K_h=H/E\in\{1,2,4,6,8,10\}$.
We use CEM with $N=512$ and stop each local solve after three consecutive
iterations without a strict decrease in its best cost.

\subsubsection{Fitting the Solver and Horizon Error Models}
\label{app:component-fits}

We fit the curves in Fig.~\ref{fig:compute-allocation}(b,d) separately.
Each fit minimizes the mean squared log ratio between predicted and
measured cost, with equal weight per configuration.

For the solver sweep, at fixed $(H,E)=(40,20)$, the full surrogate reduces
to the six-parameter form
\begin{equation}
    \widehat J_S(I,N)
    =c+\left(aI^{-u}+bN^{-v}\right)^g.
    \label{eq:app-solver-component}
\end{equation}
Here $c$ absorbs the fixed horizon contribution, $g=\eta/40$, and the
positive coefficients $a,b$ absorb the constant factor $\gamma^{40}$.
We fit all six parameters jointly to the 59 CEM configuration means.

For the horizon sweep, we fit $J_\infty+\epsilon_{\mathrm{horizon}}(H,E)$
from Eq.~\ref{eq:execution-cost-model} to the 24 configuration means,
using the zero-solver-error approximation.
Table~\ref{tab:component-fit-values} gives the fitted parameters.

For each fitting dataset $\mathcal D$, we report the mean absolute relative error as
\begin{equation}
    \operatorname{MARE}
    =\frac{100\%}{|\mathcal D|}
      \sum_{d\in\mathcal D}\left|\frac{\widehat J_d-J_d}{J_d}\right|.
    \label{eq:app-mare}
\end{equation}

\begin{table}[t]
    \centering
    \caption{Parameters of the separate component fits in
    Fig.~\ref{fig:compute-allocation}(b,d), rounded to five significant figures for display.}
    \label{tab:component-fit-values}
    \small
    \begin{tabular}{@{}lrlr@{}}
        \toprule
        Solver parameter & Value & Horizon parameter & Value \\
        \midrule
        $c$ & $0.036137$ & $J_\infty$ & $0.035463$ \\
        $a$ & $0.44596$ & $C_H$ & $1.0288$ \\
        $b$ & $0.58164$ & $\alpha_H$ & $0.10719$ \\
        $u$ & $0.19475$ & $C_E$ & $0.00042258$ \\
        $v$ & $0.35423$ & & \\
        $g$ & $3.9471$ & & \\
        \bottomrule
    \end{tabular}
\end{table}

\subsection{The Cost Model Estimates a Test-Time Compute Strategy (Q3)}
\label{app:compute-strategy-evaluation}

This section gives the fitting and evaluation details for
Sec.~\ref{sec:compute-optimal-validation}. The two evaluations report the
costs and allocation regret underlying
Figs.~\ref{fig:compute-strategy} and~\ref{fig:compute-generalization},
followed by additional analysis of configuration rankings.
We write $J(\xi)$ for the measured estimate of
$J_{\mathrm{TTC}}(\mathrm{CEM},\xi)$ and $\widehat J(\xi)$ for the proposed
surrogate's prediction, with $\xi=(I,N,H,E)$.

\subsubsection{Fitting the Surrogate Cost Model and Selecting Configurations}
\label{app:strategy-data}
\label{app:strategy-fitting}
\label{app:strategy-budget-protocol}

\noindent\textbf{Fitting data.}\ We use the motion set and cost aggregation
from Appendix~\ref{app:compute-allocation-effects}.

We fit the proposed cost model and the three regression baselines to
the same 183 equally weighted training points:
24 horizon configurations and 59 fixed-horizon CEM configurations from
Sec.~\ref{sec:compute-configuration-effects}, and 100 joint configurations
that vary all four parameters.

The joint configurations and subsequent allocation search use the integer
domain $\mathcal X_{\mathrm{exp}}$:
\begin{equation}
    \begin{gathered}
        1\le I\le512,\quad N\ge2,\quad256\le IN\le8192,\\
        5\le E\le40,\quad E\le H\le10E,\quad NH\le327680,
        \qquad I,N,H,E\in\mathbb N_+.
    \end{gathered}
    \label{eq:app-configuration-domain}
\end{equation}

To generate joint configurations, we draw $i,o,e,r$ independently from
log-uniform distributions on
$[1,512]$, $[256,8192]$, $[5,40]$, and $[1,10]$, respectively.
Here $o$ represents $IN$ and $r$ represents $H/E$.
We reject a draw if $i\ge o$. With the rounding rule
$R(x)=\lfloor x+1/2\rfloor$, set
\begin{equation}
    \begin{aligned}
        I&=R(i),\qquad E=R(e),\qquad
        H=\operatorname{clip}(R(Er),E,10E),\\
        N_{\min}&=\max\bigl(2,\lceil256/I\rceil\bigr),\\
        N_{\max}&=\min\bigl(\lfloor8192/I\rfloor,
                                \lfloor327680/H\rfloor\bigr),\\
        N&=\operatorname{clip}\bigl(R(o/I),N_{\min},N_{\max}\bigr).
    \end{aligned}
\end{equation}
The clip operation limits its first argument to the specified interval.
We reject empty intervals and tuples outside the domain.
We reject duplicate tuples and previously tested configurations.

\noindent\textbf{Proposed cost model.}\ We fit the ten positive CEM parameters
$\theta$ of the surrogate in Eq.~\ref{eq:cost-model-decomposition}.
Write $\widehat J_\theta(\xi)$ for its full prediction and
$B_\theta(H,E)=J_\infty+\epsilon_{\mathrm{horizon}}(H,E)$ for its
horizon-only prediction. Denote the horizon, solver and joint training sets by
$\mathcal D_H$, $\mathcal D_S$ and $\mathcal D_C$, respectively.
The fitting objective is
\begin{equation}
    \begin{aligned}
        \mathcal L(\theta)=\frac{1}{183}\Bigg[
        &\sum_{d\in\mathcal D_H}
          \left(\log\frac{B_\theta(H_d,E_d)}{J_d}\right)^2\\
        +&\sum_{d\in\mathcal D_S}
          \left(\log\frac{\widehat J_\theta(I_d,N_d,40,20)}{J_d}\right)^2\\
        +&\sum_{d\in\mathcal D_C}
          \left(\log\frac{\widehat J_\theta(\xi_d)}{J_d}\right)^2\Bigg].
    \end{aligned}
    \label{eq:app-joint-fit-loss}
\end{equation}
The horizon data use the zero-solver-error approximation.

We first fit the horizon parameters using the five initial values in
Table~\ref{tab:strategy-fit-starts}. We combine the best horizon fit with
each of the five solver initial values in that table and each of the twelve
$(\eta,\gamma)$ pairs in
\begin{equation}
    \bigl(\{5,20\}\times\{0.98,1\}\bigr)
    \cup\bigl(\{40,80\}\times\{0.98,1,1.02\}\bigr)
    \cup\bigl(\{140\}\times\{1,1.02\}\bigr).
\end{equation}
These combinations give 60 starts for joint optimization of all ten parameters.
We optimize the parameters in natural-log coordinates by bounded nonlinear
least squares, using the bounds in Table~\ref{tab:strategy-parameters}.
We retain the converged fit with the lowest
objective; Table~\ref{tab:strategy-parameters} lists its parameters.

\begin{table}[t]
    \centering
    \caption{Initial values for the horizon fit and the solver coordinates of the joint fit.
    Each column lists five values in the stated parameter order.}
    \label{tab:strategy-fit-starts}
    \small
    \begin{tabular}{@{}cc@{}}
        \toprule
        $(J_\infty,C_H,\alpha_H,C_E)$ & $(C_I,C_N,\alpha_I,\alpha_N)$ \\
        \midrule
        $(0.02,0.5,0.10,0.0001)$ & $(0.03,0.10,0.10,0.30)$ \\
        $(0.02,2,0.20,0.0005)$ & $(0.03,0.25,0.30,0.60)$ \\
        $(0.035,0.5,0.10,0.0001)$ & $(0.10,0.10,0.60,0.30)$ \\
        $(0.035,2,0.20,0.0005)$ & $(0.10,0.25,0.30,0.60)$ \\
        $(0.04,1,0.30,0.001)$ & $(0.25,0.25,0.60,0.60)$ \\
        \bottomrule
    \end{tabular}
\end{table}

\begin{table}[t]
    \centering
    \caption{Parameter bounds and fitted values of the complete CEM surrogate.
    Fitted values are rounded to five significant figures.}
    \label{tab:strategy-parameters}
    \small
    \begin{tabular}{@{}lcr@{}}
        \toprule
        Parameter & Bounds & Fitted value \\
        \midrule
        $J_\infty$ & $[10^{-8},1]$ & 0.034123 \\
        $C_H$ & $[10^{-8},20]$ & 1.8011 \\
        $\alpha_H$ & $[10^{-4},2]$ & 0.094838 \\
        $C_E$ & $[10^{-8},0.1]$ & 0.00047245 \\
        $C_I$ & $[10^{-8},2]$ & 0.49438 \\
        $C_N$ & $[10^{-8},2]$ & 0.83919 \\
        $\alpha_I$ & $[0.005,8]$ & 0.40234 \\
        $\alpha_N$ & $[0.005,8]$ & 0.34379 \\
        $\eta$ & $[0.1,160]$ & 160.00 \\
        $\gamma$ & $[0.9,1.1]$ & 0.98201 \\
        \bottomrule
    \end{tabular}
\end{table}

\noindent\textbf{Configuration selection.}\ We evaluate budgets
$\mathcal K=\{256,512,1024,2048,4096,8192\}$.
Table~\ref{tab:strategy-configurations} lists the six proposed allocations
and their actual compute ratios $INH/E$.

\begin{table}[t]
    \centering
    \caption{CEM allocations selected by the surrogate cost model for
    Sec.~\ref{sec:compute-optimal-validation}.}
    \label{tab:strategy-configurations}
    \small
    \begin{tabular}{@{}rrrrrrr@{}}
        \toprule
        Budget $K$ & $I$ & $N$ & $H$ & $E$ & Actual $INH/E$ & $\widehat J$ \\
        \midrule
        256  & 8  & 32 & 40 & 40 & 256  & 0.11652350 \\
        512  & 8  & 32 & 36 & 18 & 512  & 0.07089407 \\
        1024 & 9  & 31 & 33 & 9  & 1023 & 0.05898193 \\
        2048 & 10 & 48 & 34 & 8  & 2040 & 0.05236377 \\
        4096 & 13 & 54 & 35 & 6  & 4095 & 0.04772997 \\
        8192 & 15 & 78 & 35 & 5  & 8190 & 0.04439349 \\
        \bottomrule
    \end{tabular}
\end{table}

\subsubsection{Baselines}
\label{app:surrogate-baselines}

We compare the proposed cost model with random forest, radial basis function,
and Gaussian process regression. All three baselines use the fitting data in
Appendix~\ref{app:strategy-data}. We select baseline hyperparameters by mean
squared prediction error in natural-log cost on a separate validation set
of 50 joint configurations.

Each baseline maps
$x=(1/I,1/N,H,E)$ to the natural logarithm of closed-loop cost.
We linearly scale the input using fixed bounds
$l=(0,0,5,5)$ and $u=(1,1/2,400,40)$, so
$\widetilde x_j=2(x_j-l_j)/(u_j-l_j)-1$.
For the 24 points from the horizon sweep, we set the first two unscaled coordinates to
zero, representing $I,N\to\infty$ and zero solver error. 

\noindent\textbf{Random forest (RF).}\ Following AutoMPC~\citep{edwards2021autompc},
we use a regression forest with bootstrap sampling and
squared-error splits. We test 27 combinations of tree count
$\{10,30,100\}$, features per split $\{4,3,2\}$, and minimum leaf size
$\{1,3,5\}$. The minimum split size is 2.
Each leaf averages natural-log costs directly. Validation selects
10 trees, three features per split, and leaf size 1.

\noindent\textbf{Radial basis functions (RBF).}\ We follow the GLIS surrogate
used for embedded MPC calibration~\citep{forgione2020calibration}.
It applies the inverse-quadratic kernel
$\phi(r)=(1+\epsilon^2r^2)^{-1}$ to scaled inputs. We compute its weights
by truncated singular value decomposition, with no polynomial tail.
The 18 candidates combine $\epsilon\in\{0.125,0.25,0.5,1,2,4\}$
with singular-value thresholds $\{10^{-6},10^{-8},10^{-10}\}$.
Validation selects $\epsilon=0.125$ and threshold $10^{-6}$.

\noindent\textbf{Gaussian process (GP).}\ We fit exact GP regression with
a learned constant mean and Gaussian observation noise. We test
Mat\'ern $5/2$ and squared-exponential kernels, each with either a shared
length scale or a separate length scale per input coordinate.
Three initial length scales $\{0.25,1,4\}$ and three initial noise
standard deviations $\{0.001,0.03,0.3\}$ give 36 candidates.
We optimize training marginal likelihood with 250 Adam updates at learning
rate 0.05, retaining the iterate with the highest likelihood.
The log length scale, log signal standard deviation, and log noise
standard deviation are bounded by $[-6,6]$, $[-8,8]$, and $[-12,2]$,
respectively; diagonal jitter is $10^{-10}$.
Validation selects
the Mat\'ern $5/2$ kernel with separate length scales, initialized at
length 0.25 and noise standard deviation 0.3.

\noindent\textbf{Configuration selection.}\ For each baseline, we minimize
predicted log cost over the domain in Eq.~\ref{eq:app-configuration-domain}
subject to $INH/E\le K$, using the same six budgets as the proposed model.
We evaluate every feasible $(I,N,H,E)$ configuration. Table~\ref{tab:surrogate-baseline-allocations} lists the selected configurations.

\begin{table}[t]
    \centering
    \caption{Configurations $(I,N,H,E)$ selected by each regression baseline.
    The same configuration applies to both motion sets and all BFMs at
    a given budget. The new-motion evaluation uses the first five budgets.}
    \label{tab:surrogate-baseline-allocations}
    \small
    \setlength{\tabcolsep}{10pt}
    \begin{tabular}{@{}rccc@{}}
\toprule
$K$ & Random forest & RBF & Gaussian process \\
\midrule
256 & $(8,32,35,35)$ & $(128,2,40,40)$ & $(8,32,40,40)$ \\
512 & $(8,62,35,34)$ & $(256,2,40,40)$ & $(8,32,80,40)$ \\
1024 & $(14,71,35,34)$ & $(512,2,40,40)$ & $(14,37,79,40)$ \\
2048 & $(15,78,35,20)$ & $(512,2,40,40)$ & $(6,46,191,26)$ \\
4096 & $(18,130,35,20)$ & $(512,2,40,40)$ & $(7,73,191,24)$ \\
8192 & $(30,140,39,20)$ & $(512,2,40,40)$ & $(16,105,39,8)$ \\
\bottomrule
\end{tabular}

\end{table}

\subsubsection{Evaluation Metrics}
\label{app:strategy-metrics}

\noindent\textbf{Allocation regret.}\ The four strategies are
$\mathcal M=\{\mathrm{Ours},\mathrm{RF},\mathrm{RBF},\mathrm{GP}\}$.
Let $J_a(K)$ denote strategy $a$'s measured mean cost and $\mathcal R_K$
the random configurations at budget $K$ on the fitting motions.
Define the empirical optimum and proposed cost by
\begin{equation}
    J_{\mathrm{best}}(K)=\min\left\{
      \min_{\xi\in\mathcal R_K}J(\xi),\min_{a\in\mathcal M}J_a(K)\right\},
    \qquad J_{\mathrm{prop}}(K)=J(\widehat\xi^\star(K)).
\end{equation}
The relative allocation regret is
\begin{equation}
    R_K=\frac{J_{\mathrm{prop}}(K)-J_{\mathrm{best}}(K)}
                 {J_{\mathrm{best}}(K)}.
    \label{eq:app-allocation-regret}
\end{equation}
We report regret as a percentage. This is the comparison used in
Sec.~\ref{sec:compute-optimal-validation}; the baseline regrets replace
$J_{\mathrm{prop}}(K)$ with $J_a(K)$ in Eq.~\ref{eq:app-allocation-regret}.

On the five new motions, we compute the same regret separately for each
BFM $b$, using 50 random configurations and the four strategies as the
reference set. We denote this regret by $R_b(K)$.
When strategies select the same tuple, each retains its measured mean;
the Empirical Optimum error bar uses the entry with the lowest mean.

Mean regret weights budgets equally: six on the fitting motions and five
on the new motions. The overall generalization mean also weights the three
BFMs equally. Maximum regret is the largest regret across the corresponding
budgets.

\noindent\textbf{Additional analysis: configuration rankings.}\ We also
measure agreement between predicted and measured costs across the random
configurations, using Kendall's $\tau_b$, Spearman's $\rho$, and NDCG@5.
For the fitting-motion evaluation, these metrics use the $n=200$
configurations in $\mathcal R_K$ at each budget. Their measured and predicted
costs are $y_i=J(\xi_i)$ and $\widehat y_i=\widehat J(\xi_i)$, with ascending
ranks $r_i$ and $\widehat r_i$, respectively. Rank one denotes the lowest cost.
Tied costs receive their average rank. All costs in each measured or
predicted vector are distinct in this experiment.

\noindent\textbf{Kendall's $\tau_b$.}\ A pair is concordant if both orderings
agree on which cost is lower and discordant if they disagree. Let $n_c$ and
$n_d$ count these pairs, $n_0=n(n-1)/2$, and let $t_y$ and $t_{\widehat y}$
count pairs tied in measured and predicted cost. Then
\begin{equation}
    \tau_b=\frac{n_c-n_d}
    {\sqrt{(n_0-t_y)(n_0-t_{\widehat y})}}.
    \label{eq:app-kendall}
\end{equation}
Pairs tied in both costs contribute to both tie counts. For distinct costs,
$\tau_b=(n_c-n_d)/n_0$~\citep{Kendall}. A value of one means complete agreement,
zero means equal numbers of concordant and discordant pairs, and minus one
means reverse order.
For these distinct costs, the fraction of correctly ordered
pairs is $n_c/n_0=(1+\tau_b)/2$.

\noindent\textbf{Spearman's $\rho$.}\ Spearman correlation~\citep{Spearman} is the Pearson
correlation between the two rank vectors:
\begin{equation}
    \rho=\frac{\sum_{i=1}^{n}(r_i-\bar r)(\widehat r_i-\overline{\widehat r})}
    {\sqrt{\sum_{i=1}^{n}(r_i-\bar r)^2}
     \sqrt{\sum_{i=1}^{n}(\widehat r_i-\overline{\widehat r})^2}}.
    \label{eq:app-spearman}
\end{equation}
The bars denote the means of the rank vectors. A value of one means matching
ranks, while minus one means reverse ranks.

\noindent\textbf{NDCG@5~\citep{NDCG}.}\ To assess the top of the predicted ranking,
assign linear relevance $g_i=n+1-r_i=201-r_i$.
Let $\widehat\pi(j)$ denote the configuration at predicted position $j$,
and $\pi(j)$ the configuration at measured position $j$. Then
\begin{equation}
    \begin{aligned}
        \mathrm{DCG@5}
        &=\sum_{j=1}^{5}\frac{g_{\widehat\pi(j)}}{\log_2(j+1)},
        &\mathrm{IDCG@5}
        &=\sum_{j=1}^{5}\frac{g_{\pi(j)}}{\log_2(j+1)},\\
        \mathrm{NDCG@5}&=\frac{\mathrm{DCG@5}}{\mathrm{IDCG@5}}.
    \end{aligned}
    \label{eq:app-ndcg}
\end{equation}
Higher relevance means lower measured cost. The discount gives more weight
to earlier predicted positions, and the denominator is the ideal discounted
gain. A value of one attains the ideal top-five ranking under this gain.

On the five new motions, ranking metrics use the 50 random configurations
per BFM and budget, with predictions from the SONIC-fitted surrogate.
With $n=50$, NDCG@5 relevance becomes $g_i=n+1-r_i=51-r_i$.
The tie rules are unchanged.

We compute each ranking metric within a budget, then average equally over
the six budgets on the fitting motions or the five budgets for each BFM
on the new motions.

\subsubsection{Compute Scaling on the Fitting Motions}
\label{app:strategy-fitting-evaluation}

\noindent\textbf{Evaluation protocol.}\ We evaluate the four models'
selected configurations on the ten SONIC fitting motions
(Table~\ref{tab:allocation-motions}) at each of the six budgets in $\mathcal K$.
The comparison uses three seeds and 200 distinct random configurations per
budget, each satisfying $0.9K\le INH/E\le K$.

For $K>256$, we sample configurations using the procedure in
Appendix~\ref{app:strategy-fitting}, subject to the budget band and without
repeating configurations across budgets. Previously tested configurations
remain eligible. At $K=256$, we sample without replacement from feasible
configurations with $IN=256$ and $H=E$, using the probabilities induced by
the same continuous sampling and rounding procedure.

\noindent\textbf{Costs and regret in the main comparison.}\
Tables~\ref{tab:surrogate-baseline-results} and~\ref{tab:surrogate-baseline-regret}
give the costs and regret for the four strategies compared in
Fig.~\ref{fig:compute-strategy} and Sec.~\ref{sec:compute-optimal-validation}.

\begin{table}[htbp]
    \centering
    \caption{Measured costs on the ten SONIC fitting motions at matched
    compute budgets; lower is better.
    Entries report mean cost $\pm$ one sample standard deviation across
    the three seed means. Empirical Optimum uses the configuration with
    the lowest mean cost among the four strategies and 200 random configurations.}
    \label{tab:surrogate-baseline-results}
    \footnotesize
    \setlength{\tabcolsep}{2pt}
    \begin{tabular}{@{}rrrrrr@{}}
\toprule
$K$ & Ours & Random forest & RBF & Gaussian process & Empirical Optimum \\
\midrule
\multicolumn{6}{l}{\emph{Ten fitting motions, SONIC}} \\
256 & $0.1048\pm0.0063$ & $0.1226\pm0.0310$ & $0.2096\pm0.0260$ & $0.1053\pm0.0068$ & $0.1027\pm0.0028$ \\
512 & $0.0715\pm0.0034$ & $0.0923\pm0.0073$ & $0.1554\pm0.0360$ & $0.0855\pm0.0140$ & $0.0715\pm0.0034$ \\
1024 & $0.0607\pm0.0052$ & $0.0783\pm0.0141$ & $0.1461\pm0.0350$ & $0.0799\pm0.0065$ & $0.0570\pm0.0017$ \\
2048 & $0.0557\pm0.0021$ & $0.0585\pm0.0093$ & $0.1461\pm0.0350$ & $0.0868\pm0.0110$ & $0.0528\pm0.0049$ \\
4096 & $0.0484\pm0.0010$ & $0.0520\pm0.0049$ & $0.1461\pm0.0350$ & $0.0741\pm0.0015$ & $0.0461\pm0.0028$ \\
8192 & $0.0441\pm0.0007$ & $0.0425\pm0.0005$ & $0.1461\pm0.0350$ & $0.0419\pm0.0009$ & $0.0419\pm0.0009$ \\
\bottomrule
\end{tabular}

\end{table}

\begin{table}[htbp]
    \centering
    \caption{Allocation regret (\%) on the ten SONIC fitting motions.
    The reference is Empirical Optimum in Table~\ref{tab:surrogate-baseline-results}.
    The Mean row weights the six budgets equally. Bold marks the lowest
    regret in each row.}
    \label{tab:surrogate-baseline-regret}
    \small
    \setlength{\tabcolsep}{8pt}
    \begin{tabular}{@{}llrrrr@{}}
\toprule
BFM & $K$ & Ours & Random forest & RBF & Gaussian process \\
\midrule
\multicolumn{6}{l}{\emph{Figure~\ref{fig:compute-strategy}: ten fitting motions}} \\
SONIC & 256 & \textbf{2.10} & 19.36 & 104.13 & 2.60 \\
 & 512 & \textbf{0.00} & 29.08 & 117.26 & 19.52 \\
 & 1024 & \textbf{6.40} & 37.27 & 156.16 & 40.12 \\
 & 2048 & \textbf{5.38} & 10.65 & 176.43 & 64.25 \\
 & 4096 & \textbf{5.15} & 12.90 & 217.10 & 60.90 \\
 & 8192 & 5.10 & 1.35 & 248.25 & \textbf{0.00} \\
\cmidrule(l){2-6}
 & Mean & \textbf{4.02} & 18.44 & 169.89 & 31.23 \\
\bottomrule
\end{tabular}

\end{table}

\noindent\textbf{Additional analysis: configuration rankings.}\
Table~\ref{tab:compute-strategy} reports the proposed model's ranking
accuracy on the random configurations at each budget.

\begin{table}[!htbp]
    \centering
    \caption{\textbf{Additional ranking analysis in the SONIC scaling experiment.}
    Ranking metrics use the 200 random configurations at each budget;
    regret refers to the proposed configuration. The Mean row weights
    the six budgets equally.}
    \label{tab:compute-strategy}
    \small
    \setlength{\tabcolsep}{8pt}
    \begin{tabular}{@{}rrrrr@{}}
\toprule
$K$ & $\tau\uparrow$ & $\rho\uparrow$ & NDCG@5$\uparrow$ & Regret (\%)$\downarrow$ \\
\midrule
256 & 0.722 & 0.901 & 0.954 & 2.10 \\
512 & 0.731 & 0.902 & 0.987 & 0.00 \\
1024 & 0.754 & 0.917 & 0.935 & 6.40 \\
2048 & 0.562 & 0.730 & 0.798 & 5.38 \\
4096 & 0.552 & 0.711 & 0.981 & 5.15 \\
8192 & 0.572 & 0.736 & 0.944 & 5.10 \\
\midrule
Mean & 0.649 & 0.816 & 0.933 & 4.02 \\
\bottomrule
\end{tabular}

\end{table}

\newpage
\subsubsection{Generalization to New Motions and BFMs}
\label{app:compute-generalization}

\noindent\textbf{Evaluation protocol.}\ Figure~\ref{fig:compute-generalization}
compares the four strategies on SONIC, HEFT, and TWIST2 using five AMASS
motions disjoint from the fitting set (Table~\ref{tab:generalization-motions}).
This experiment uses $\mathcal K_{\mathrm{test}}=\{256,512,1024,2048,4096\}$.
Each model uses its first five allocations from
Tables~\ref{tab:strategy-configurations} and~\ref{tab:surrogate-baseline-allocations}
on all three BFMs.

At each budget, all three BFMs use the same 50 distinct random
configurations from the fitting-motion experiment. We evaluate these and
the selected configurations on the new motions with three optimizer seeds,
using the 14 bodies listed in Appendix~\ref{app:large-scale-motion-tracking}.

We reuse measurements when a configuration is selected at several budgets
for the same BFM and motion set. RBF selects
$(512,2,40,40)$ for every $K\ge1024$, using actual compute 1024 and
producing a flat cost curve.

\begin{table}[htbp]
    \centering
    \caption{New motions used to evaluate strategy generalization.
    Durations count executed control steps at $0.02$\,s per step.}
    \label{tab:generalization-motions}
    \footnotesize
    \setlength{\tabcolsep}{4pt}
    \begin{tabular}{@{}lrr@{}}
        \toprule
        Source sequence & Frames & Duration (s) \\
        \midrule
        \texttt{CMU/143/143\_14\_poses} & 617 & 12.32 \\
        \texttt{BMLmovi/Subject\_57\_F\_MoSh/Subject\_57\_F\_2\_poses} & 277 & 5.52 \\
        \texttt{EKUT/300/PushBeamPedar\_15\_poses} & 247 & 4.92 \\
        \texttt{BioMotionLab\_NTroje/rub012/0009\_normal\_jog1\_poses} & 109 & 2.16 \\
        \texttt{CMU/74/74\_01\_poses} & 400 & 7.98 \\
        \bottomrule
    \end{tabular}
\end{table}

The nominal-command costs labeled in Fig.~\ref{fig:compute-generalization}
average the paired baseline executions equally over the five motions and
three seeds.

\noindent\textbf{Costs and regret in the main comparison.}\
Tables~\ref{tab:surrogate-baseline-generalization-results}
and~\ref{tab:surrogate-baseline-generalization-regret} give the costs and
regret for all four strategies in Fig.~\ref{fig:compute-generalization}.

\begin{table}[tp]
    \centering
    \caption{Measured costs on five new motions at matched compute budgets;
    lower is better. Entries report mean cost $\pm$ one sample standard
    deviation across the three seed means. Empirical Optimum uses the
    configuration with the lowest mean cost among the four strategies and
    50 random configurations on each BFM.}
    \label{tab:surrogate-baseline-generalization-results}
    \footnotesize
    \setlength{\tabcolsep}{2pt}
    \begin{tabular}{@{}lrrrrrr@{}}
\toprule
BFM & $K$ & Ours & Random forest & RBF & Gaussian process & Empirical Optimum \\
\midrule
\multicolumn{7}{l}{\emph{Five new motions, SONIC}} \\
SONIC & 256 & $0.1621\pm0.0166$ & $0.1553\pm0.0318$ & $0.2653\pm0.0167$ & $0.1617\pm0.0171$ & $0.1553\pm0.0318$ \\
SONIC & 512 & $0.0875\pm0.0040$ & $0.1174\pm0.0104$ & $0.2346\pm0.0225$ & $0.0786\pm0.0053$ & $0.0752\pm0.0045$ \\
SONIC & 1024 & $0.0660\pm0.0026$ & $0.0887\pm0.0065$ & $0.2014\pm0.0141$ & $0.0722\pm0.0028$ & $0.0647\pm0.0033$ \\
SONIC & 2048 & $0.0559\pm0.0004$ & $0.0728\pm0.0067$ & $0.2014\pm0.0141$ & $0.0849\pm0.0035$ & $0.0559\pm0.0004$ \\
SONIC & 4096 & $0.0542\pm0.0011$ & $0.0603\pm0.0023$ & $0.2014\pm0.0141$ & $0.0827\pm0.0016$ & $0.0537\pm0.0022$ \\
\midrule
\multicolumn{7}{l}{\emph{Five new motions, HEFT}} \\
HEFT & 256 & $0.0560\pm0.0015$ & $0.0516\pm0.0002$ & $0.0755\pm0.0068$ & $0.0554\pm0.0011$ & $0.0516\pm0.0002$ \\
HEFT & 512 & $0.0486\pm0.0008$ & $0.0484\pm0.0009$ & $0.0641\pm0.0033$ & $0.0575\pm0.0005$ & $0.0463\pm0.0015$ \\
HEFT & 1024 & $0.0447\pm0.0004$ & $0.0426\pm0.0006$ & $0.0593\pm0.0035$ & $0.0540\pm0.0008$ & $0.0426\pm0.0006$ \\
HEFT & 2048 & $0.0409\pm0.0005$ & $0.0374\pm0.0002$ & $0.0593\pm0.0035$ & $0.0689\pm0.0006$ & $0.0374\pm0.0002$ \\
HEFT & 4096 & $0.0385\pm0.0003$ & $0.0346\pm0.0001$ & $0.0593\pm0.0035$ & $0.0671\pm0.0009$ & $0.0340\pm0.0004$ \\
\midrule
\multicolumn{7}{l}{\emph{Five new motions, TWIST2}} \\
TWIST2 & 256 & $0.5942\pm0.0371$ & $0.6087\pm0.0446$ & $1.3601\pm0.3610$ & $0.6090\pm0.0319$ & $0.5780\pm0.0269$ \\
TWIST2 & 512 & $0.4958\pm0.0076$ & $0.5534\pm0.0432$ & $0.8815\pm0.1238$ & $0.5672\pm0.0129$ & $0.4885\pm0.0123$ \\
TWIST2 & 1024 & $0.4468\pm0.0070$ & $0.5107\pm0.0110$ & $0.8148\pm0.0421$ & $0.5377\pm0.0224$ & $0.4468\pm0.0070$ \\
TWIST2 & 2048 & $0.4296\pm0.0015$ & $0.4501\pm0.0137$ & $0.8148\pm0.0421$ & $0.6045\pm0.0062$ & $0.4296\pm0.0015$ \\
TWIST2 & 4096 & $0.4065\pm0.0048$ & $0.4228\pm0.0124$ & $0.8148\pm0.0421$ & $0.5814\pm0.0065$ & $0.4065\pm0.0048$ \\
\bottomrule
\end{tabular}

\end{table}

\begin{table}[tp]
    \centering
    \caption{Allocation regret (\%) on five new motions.
    The reference is Empirical Optimum in
    Table~\ref{tab:surrogate-baseline-generalization-results}.
    Each Mean row weights the five budgets equally; the
    Figure~\ref{fig:compute-generalization} mean also weights BFMs equally.
    Bold marks the lowest regret in each row.}
    \label{tab:surrogate-baseline-generalization-regret}
    \small
    \setlength{\tabcolsep}{8pt}
    \begin{tabular}{@{}llrrrr@{}}
\toprule
BFM & $K$ & Ours & Random forest & RBF & Gaussian process \\
\midrule
\multicolumn{6}{l}{\emph{Figure~\ref{fig:compute-generalization}: five new motions}} \\
SONIC & 256 & 4.34 & \textbf{0.00} & 70.81 & 4.07 \\
 & 512 & 16.27 & 56.07 & 211.91 & \textbf{4.52} \\
 & 1024 & \textbf{2.07} & 37.10 & 211.23 & 11.52 \\
 & 2048 & \textbf{0.00} & 30.22 & 260.27 & 51.91 \\
 & 4096 & \textbf{0.91} & 12.36 & 275.05 & 53.99 \\
\cmidrule(l){2-6}
 & Mean & \textbf{4.72} & 27.15 & 205.86 & 25.20 \\
\midrule
HEFT & 256 & 8.59 & \textbf{0.00} & 46.34 & 7.36 \\
 & 512 & 4.98 & \textbf{4.59} & 38.61 & 24.34 \\
 & 1024 & 4.77 & \textbf{0.00} & 39.16 & 26.57 \\
 & 2048 & 9.48 & \textbf{0.00} & 58.75 & 84.27 \\
 & 4096 & 13.37 & \textbf{1.94} & 74.55 & 97.28 \\
\cmidrule(l){2-6}
 & Mean & 8.24 & \textbf{1.31} & 51.48 & 47.96 \\
\midrule
TWIST2 & 256 & \textbf{2.80} & 5.31 & 135.31 & 5.36 \\
 & 512 & \textbf{1.49} & 13.29 & 80.45 & 16.12 \\
 & 1024 & \textbf{0.00} & 14.30 & 82.36 & 20.36 \\
 & 2048 & \textbf{0.00} & 4.77 & 89.64 & 40.70 \\
 & 4096 & \textbf{0.00} & 4.03 & 100.45 & 43.05 \\
\cmidrule(l){2-6}
 & Mean & \textbf{0.86} & 8.34 & 97.64 & 25.12 \\
\midrule
\multicolumn{2}{l}{Figure~\ref{fig:compute-generalization} mean} & \textbf{4.60} & 12.27 & 118.33 & 32.76 \\
\bottomrule
\end{tabular}

\end{table}

\clearpage
\noindent\textbf{Additional analysis: configuration rankings.}\
Table~\ref{tab:compute-generalization} reports the SONIC-fitted surrogate's
ranking accuracy on the random configurations for each BFM and budget.

\begin{table}[!htbp]
    \centering
    \caption{\textbf{Additional ranking analysis on new motions and BFMs.}
    Costs average five motions and three seeds.
    Ranking metrics use the 50 random configurations per budget;
    regret refers to the proposed configuration.
    Each Mean row averages five budgets, and each Max row gives the
    largest regret among those budgets.}
    \label{tab:compute-generalization}
    \small
    \setlength{\tabcolsep}{4.5pt}
    \begin{tabular}{@{}lrrrrrrr@{}}
\toprule
BFM & $K$ & $J_{\mathrm{prop}}\downarrow$ & $J_{\mathrm{best}}\downarrow$ & Regret (\%)$\downarrow$ & $\tau\uparrow$ & $\rho\uparrow$ & NDCG@5$\uparrow$ \\
\midrule
SONIC & 256 & 0.1621 & 0.1553 & 4.34 & 0.726 & 0.886 & 0.985 \\
SONIC & 512 & 0.0875 & 0.0752 & 16.27 & 0.758 & 0.916 & 0.956 \\
SONIC & 1024 & 0.0660 & 0.0647 & 2.07 & 0.678 & 0.842 & 0.902 \\
SONIC & 2048 & 0.0559 & 0.0559 & 0.00 & 0.479 & 0.631 & 0.691 \\
SONIC & 4096 & 0.0542 & 0.0537 & 0.91 & 0.527 & 0.703 & 0.945 \\
\cmidrule(lr){2-8}
SONIC & Mean & 0.0851 & 0.0810 & 4.72 & 0.634 & 0.795 & 0.896 \\
SONIC & Max & --- & --- & 16.27 & --- & --- & --- \\
\midrule
HEFT & 256 & 0.0560 & 0.0516 & 8.59 & 0.806 & 0.944 & 0.945 \\
HEFT & 512 & 0.0486 & 0.0463 & 4.98 & 0.729 & 0.891 & 0.974 \\
HEFT & 1024 & 0.0447 & 0.0426 & 4.77 & 0.154 & 0.223 & 0.691 \\
HEFT & 2048 & 0.0409 & 0.0374 & 9.48 & 0.265 & 0.375 & 0.566 \\
HEFT & 4096 & 0.0385 & 0.0340 & 13.37 & 0.360 & 0.500 & 0.909 \\
\cmidrule(lr){2-8}
HEFT & Mean & 0.0457 & 0.0424 & 8.24 & 0.463 & 0.586 & 0.817 \\
HEFT & Max & --- & --- & 13.37 & --- & --- & --- \\
\midrule
TWIST2 & 256 & 0.5942 & 0.5780 & 2.80 & 0.615 & 0.804 & 0.946 \\
TWIST2 & 512 & 0.4958 & 0.4885 & 1.49 & 0.758 & 0.919 & 0.983 \\
TWIST2 & 1024 & 0.4468 & 0.4468 & 0.00 & 0.675 & 0.839 & 0.897 \\
TWIST2 & 2048 & 0.4296 & 0.4296 & 0.00 & 0.448 & 0.595 & 0.601 \\
TWIST2 & 4096 & 0.4065 & 0.4065 & 0.00 & 0.456 & 0.611 & 0.941 \\
\cmidrule(lr){2-8}
TWIST2 & Mean & 0.4746 & 0.4699 & 0.86 & 0.591 & 0.754 & 0.874 \\
TWIST2 & Max & --- & --- & 2.80 & --- & --- & --- \\
\bottomrule
\end{tabular}

\end{table}

\clearpage
\subsection{The Benefits of Test-Time Compute Extend to Loco-Manipulation (Q4)}
\label{app:downstream}

\subsubsection{Label Comparison and Controller Interfaces}
\label{app:downstream-labels}

The source BFM is used during demonstration collection; the target BFM
executes commands predicted by the learned high-level policy. In simulation,
Original and State each produce one dataset per source and task, so their
trained policies are reused across SONIC, TWIST2, and HEFT. Ours produces
a separate set of labels and a trained policy for each target BFM.

\subsubsection{Dynamics Randomization During Optimization}
\label{app:downstream-randomization}

We optimize commands in MJWarp for execution in Isaac Sim or on the
physical robot. To account for differences in dynamics, each candidate
is evaluated in $R=4$ randomized physics scenarios. The fixed CEM
configuration is $(I,N,H,E)=(7,39,37,5)$ with nominal budget $K=2048$;
including all scenarios gives a rollout budget of $RINH/E$.

At each planning update, we draw dynamics multipliers and initial-velocity
perturbations from Table~\ref{tab:downstream-randomization}. All candidates
share this scenario set across optimizer iterations, and the next planning
update draws a new set. The demonstrated reference motion stays fixed.
Each candidate's score is the arithmetic mean of its four scenario costs.

\begin{table}[htbp]
    \centering
    \caption{Randomization distributions used in downstream label optimization.
    Multipliers scale nominal dynamics parameters; velocity perturbations
    are additive, zero-mean Gaussian offsets.}
    \label{tab:downstream-randomization}
    \small
    \begin{tabular}{@{}ll@{}}
        \toprule
        Parameter & Distribution \\
        \midrule
        Contact-friction multiplier & $\mathcal U(0.8,1.2)$ \\
        Joint-armature multiplier & $\mathcal U(0.8,1.2)$ \\
        Joint-friction-loss multiplier & $\mathcal U(0.8,1.2)$ \\
        Actuation-strength multiplier & $\mathcal U(0.9,1.1)$ \\
        Root linear velocity offset & $\mathcal N(0,0.02^2)$ per component \\
        Root angular velocity offset & $\mathcal N(0,0.05^2)$ per component \\
        Joint velocity offset & $\mathcal N(0,0.05^2)$ per component \\
        \bottomrule
    \end{tabular}
\end{table}

\subsubsection{HumanoidArena Data, Training, and Evaluation}
\label{app:downstream-simulation}

We use five HumanoidArena~\citep{wang2026humanoidarena} tasks with the
officially released LeRobot demonstrations. Closed-loop evaluation runs in Isaac Sim 5.0.0
with Isaac Lab 2.2.0 and Unitree Dex3 hands. Object randomization, lighting,
and textures follow the collection conditions. All target BFMs operate at
50 Hz with frozen pretrained parameters. 

\noindent\textbf{Command labels.}\ State encodes the recorded achieved motion
into 40-dimensional actions comprising local displacement increments, root
height, a six-dimensional orientation representation, 29 joint targets, and
two source hand actions. Every frame encodes its achieved pose, including
reset frames.

For Ours, the first 38 label dimensions encode the optimized motion commands
applied to the target BFM during optimization using the same representation.
Reset frames encode the achieved pose, and the two hand-action dimensions
retain the source labels.

\noindent\textbf{Training and evaluation.}\ We train Diffusion Policy~\citep{chi2023diffusionpolicy} with
three seeds for each condition, using the frozen source--task training
configuration. Training uses 100,000 updates with batch size 64, a ResNet18
visual encoder, and Adam with learning rate $10^{-4}$. The action chunk horizon
is 24 steps, of which 20 are executed before the next policy query.
We evaluate the 100,000-step checkpoint for every seed using 60 rollouts,
for a total of 180 rollouts per source--task--target and label condition.

Each rollout starts from the task's fixed default robot pose with
randomized objects; initializations are matched across label conditions.
For the Mean column, we first average the five task success rates within
each seed, then compute the mean and sample standard deviation across seeds.

An episode ends when the task succeeds, the robot falls, or the task's
step limit is reached. Table~\ref{tab:downstream-sim-tasks} gives the success
conditions and limits.

\begin{table}[htbp]
    \centering
    \caption{Success conditions and episode limits for the five HumanoidArena tasks.}
    \label{tab:downstream-sim-tasks}
    \small
    \setlength{\tabcolsep}{4pt}
    \begin{tabular}{@{}p{0.19\linewidth}p{0.65\linewidth}r@{}}
        \toprule
        Task & Success condition & Max steps \\
        \midrule
        Boxing & The hand/fist proxy enters the target sphere around the punching bag. & 900 \\
        Open Door & The latch is unlocked and the door opens at least $60^\circ$, with root height $\geq0.45$ m and up-axis $z\geq0.60$. & 1,800 \\
        Pick \& Place Box & The box rests on the shelf support surface. & 1,450 \\
        Sit Sofa & Sitting-related links enter the seat region with contact, or satisfy the geometric occupancy criterion. & 2,000 \\
        Vision Navigation & Both feet reach the target region while the robot remains standing. & 1,800 \\
        \bottomrule
    \end{tabular}
\end{table}

\subsubsection{Real-Robot Data, Training, and Evaluation}
\label{app:downstream-hardware}

We use 50 SONIC-collected demonstrations per task, approximately 20 minutes
of data, to fine-tune GR00T N1.7-3B~\citep{nvidia2026gr00tn17}. The physical robot is a Unitree G1 with
a 29-DoF body, BrainCo Revo2 hands, and a ZED-M camera. SONIC execution uses
the released low-latency whole-body controller~\citep{luo2026sonic}. Table~\ref{tab:downstream-interface}
summarizes the policy interface. Each policy query uses one egocentric image,
a language task description, and proprioception to predict 40 action steps. SONIC decodes the body
tokens into motor-angle targets at approximately 50 Hz. Hand commands
follow a separate control path.

\begin{table}[htbp]
    \centering
    \caption{Real-robot observations and actions at the policy interface.
    The body-and-hand state contains 41 values, with three additional
    projected-gravity components computed from the IMU orientation.}
    \label{tab:downstream-interface}
    \small
    \begin{tabular}{@{}lll@{}}
        \toprule
        Signal & Size & Representation \\
        \midrule
        Language & Text & Task description \\
        Egocentric frame & $376\times672\times3$ & ZED-M left eye (H$\times$W$\times$C) \\
        Body state & 29 & Joint values \\
        Hand state & $6+6$ & BrainCo Revo2, normalized to $[0,1]$ \\
        Projected gravity & 3 & Computed from IMU orientation \\
        Body action & 64 & SONIC motion token \\
        Hand action & $6+6$ & BrainCo Revo2, normalized to $[0,1]$ \\
        \bottomrule
    \end{tabular}
\end{table}

\noindent\textbf{Labels and fine-tuning.}\ During demonstration collection,
we record the G1 motion references supplied to SONIC before token encoding,
together with measured joint positions and root poses obtained from robot
state estimation.

For SONIC execution, Original retains the stored collection-time SONIC
motion tokens. State encodes the recorded achieved motion into
64-dimensional SONIC tokens. Ours first exports the optimized
commands applied to the BFM during offline optimization, then encodes
these commands into motion-token labels. Temporal windows, frame
alignment, and reset handling follow the SONIC low-latency defaults.\footnote{\url{https://nvlabs.github.io/GR00T-WholeBodyControl/tutorials/vla_workflow.html}}

For HEFT execution, we use the released pretrained controller with frozen
weights. Original expresses the recorded source G1 references in HEFT's
motion-command representation. State uses the achieved motion assembled
from the estimated root poses and measured joint positions. Ours uses
commands optimized offline with the frozen HEFT controller to track the
same achieved motion. All three conditions use the same action representation
and retain the demonstration observations and hand-command labels.

The high-level policy predicts 29 joint targets and a nine-dimensional
root-motion encoding per action step, together with the same 12 hand commands.
The root encoding comprises sine and cosine-minus-one pairs for roll and
pitch, the wrapped yaw increment, three position increments in the current
yaw frame, and root height. At deployment, the adapter reconstructs root
trajectories using the maintained reference XY position and the measured
IMU heading, and supplies these trajectories and joint targets to HEFT.
The root-pose estimates recorded during collection support offline label
construction; deployment does not require external root-position odometry.

We fine-tune the action head with the vision and language backbone frozen,
using the Isaac-GR00T implementation.\footnote{\url{https://github.com/NVIDIA/Isaac-GR00T}}
Table~\ref{tab:downstream-finetuning} lists the shared fine-tuning parameters
for SONIC execution.

\begin{table}[htbp]
    \centering
    \caption{Shared GR00T fine-tuning configuration for the real-robot experiments with SONIC execution.}
    \label{tab:downstream-finetuning}
    \small
    \begin{tabular}{@{}ll@{}}
        \toprule
        Parameter & Value \\
        \midrule
        Optimizer & AdamW \\
        Learning rate & $10^{-4}$ \\
        Global batch size & 160 \\
        Learning-rate schedule & Cosine, 30,000 updates \\
        Warmup & 5\% (1,500 updates) \\
        Weight decay & $10^{-5}$ \\
        Maximum gradient norm & 1.0 \\
        State dropout & 0.2 \\
        Evaluated checkpoint & 20,000 updates \\
        \bottomrule
    \end{tabular}
\end{table}

\noindent\textbf{Tasks and success.}\ Table~\ref{tab:downstream-real-tasks}
describes the four tasks and their completion conditions.
We evaluate each checkpoint for 60 rollouts. A trial starts near the
demonstration's initial scene configuration and ends when the task succeeds
or failure is judged. 

\begin{center}
\begin{minipage}{\linewidth}
    \centering
    \captionof{table}{Real-robot task sequences and success conditions.
    Shelf placements use the designated target location.}
    \label{tab:downstream-real-tasks}
    \small
    \setlength{\tabcolsep}{4pt}
    \begin{tabular}{@{}p{0.18\linewidth}p{0.46\linewidth}p{0.30\linewidth}@{}}
        \toprule
        Task & Execution sequence & Success condition \\
        \midrule
        Cup in basket & Approach the table with one or two short steps, grasp the cup with the right hand, and place it in the basket to its right. & The cup is picked up and placed in the basket. \\
        Parcel bag on shelf & Grasp the bag from the table, move and turn left toward the shelf, and place it at the target location. & The bag is picked up and placed at the target shelf location. \\
        Basket on shelf & Squat to grasp the low basket with the right hand, rise while lifting it, then move and turn left to place it on the shelf. & The basket is picked up and placed at the target shelf location. \\
        Bottle in drawer & Grasp the bottle from the cabinet top, lower it into the initially open drawer, and push the drawer closed. & The bottle is inside the drawer and the drawer is closed. \\
        \bottomrule
    \end{tabular}
\end{minipage}
\end{center}

\subsubsection{Complete Simulation Results}
\label{app:downstream-results}

Table~\ref{tab:downstream-full} reports the complete five-task comparison.
The Mean column equally weights the five tasks.
The Mean column is computed before rounding task success rates.

\begin{table}[htbp]
    \centering
    \caption{HumanoidArena task success rates (\%) for all evaluated source--target
    pairings in the five-task subset. Source is the demonstration-collection
    BFM and target is the execution BFM. Original, State, and Ours denote
    the three label conditions. Results are reported as mean $\pm$ sample
    standard deviation across three training seeds, with 60 evaluation
    rollouts per task and seed.}
    \label{tab:downstream-full}
    \fontsize{9}{10}\selectfont
    \setlength{\tabcolsep}{2.5pt}
    \begin{tabular}{@{}lllrrrrrr@{}}
\toprule
Source & Target & Labels & Boxing & \shortstack{Open\\Door} & \shortstack{Pick \&\\Place Box} & \shortstack{Sit\\Sofa} & \shortstack{Vision\\Navigation} & Mean \\
\midrule
SONIC & SONIC & Original & $83.9\pm1.0$ & $84.4\pm1.9$ & $61.1\pm1.0$ & $88.9\pm1.0$ & $60.6\pm1.0$ & $75.8\pm0.5$ \\
SONIC & SONIC & State & $0.0\pm0.0$ & $0.0\pm0.0$ & $0.0\pm0.0$ & $0.0\pm0.0$ & $0.0\pm0.0$ & $0.0\pm0.0$ \\
SONIC & SONIC & Ours & $63.9\pm1.0$ & $51.1\pm1.0$ & $54.4\pm1.0$ & $66.7\pm1.7$ & $47.2\pm1.0$ & $56.7\pm0.3$ \\
\midrule
SONIC & TWIST2 & Original & $48.9\pm1.9$ & $0.0\pm0.0$ & $0.0\pm0.0$ & $17.8\pm1.9$ & $11.7\pm1.7$ & $15.7\pm0.6$ \\
SONIC & TWIST2 & State & $33.9\pm1.0$ & $0.0\pm0.0$ & $0.0\pm0.0$ & $14.4\pm1.0$ & $1.7\pm1.7$ & $10.0\pm0.3$ \\
SONIC & TWIST2 & Ours & $50.0\pm1.7$ & $0.0\pm0.0$ & $0.0\pm0.0$ & $21.7\pm1.7$ & $7.2\pm1.9$ & $15.8\pm0.4$ \\
\midrule
SONIC & HEFT & Original & $46.7\pm1.7$ & $0.0\pm0.0$ & $0.0\pm0.0$ & $22.8\pm1.9$ & $16.1\pm1.0$ & $17.1\pm0.4$ \\
SONIC & HEFT & State & $38.9\pm1.9$ & $0.0\pm0.0$ & $0.0\pm0.0$ & $36.1\pm1.0$ & $27.8\pm1.9$ & $20.6\pm0.7$ \\
SONIC & HEFT & Ours & $57.2\pm1.0$ & $0.0\pm0.0$ & $0.0\pm0.0$ & $67.8\pm1.0$ & $18.9\pm1.0$ & $28.8\pm0.2$ \\
\midrule
TWIST2 & TWIST2 & Original & $55.0\pm1.7$ & $71.1\pm1.0$ & $41.7\pm2.9$ & $59.4\pm1.0$ & $59.4\pm1.9$ & $57.3\pm1.2$ \\
TWIST2 & TWIST2 & State & $38.3\pm2.9$ & $0.0\pm0.0$ & $0.0\pm0.0$ & $72.8\pm1.9$ & $44.4\pm2.5$ & $31.1\pm0.4$ \\
TWIST2 & TWIST2 & Ours & $50.0\pm1.7$ & $43.3\pm1.7$ & $42.8\pm1.0$ & $48.3\pm2.9$ & $55.6\pm2.5$ & $48.0\pm1.7$ \\
\midrule
TWIST2 & SONIC & Original & $0.0\pm0.0$ & $0.0\pm0.0$ & $0.0\pm0.0$ & $2.8\pm1.9$ & $36.7\pm2.9$ & $7.9\pm0.8$ \\
TWIST2 & SONIC & State & $5.0\pm2.9$ & $0.0\pm0.0$ & $0.0\pm0.0$ & $22.8\pm2.5$ & $50.6\pm2.5$ & $15.7\pm1.2$ \\
TWIST2 & SONIC & Ours & $37.8\pm2.5$ & $0.0\pm0.0$ & $0.0\pm0.0$ & $56.7\pm2.9$ & $34.4\pm1.9$ & $25.8\pm1.4$ \\
\midrule
TWIST2 & HEFT & Original & $36.7\pm1.7$ & $0.0\pm0.0$ & $0.0\pm0.0$ & $31.1\pm1.0$ & $4.4\pm1.0$ & $14.4\pm0.2$ \\
TWIST2 & HEFT & State & $59.4\pm1.0$ & $0.0\pm0.0$ & $0.0\pm0.0$ & $46.7\pm2.9$ & $50.6\pm2.5$ & $31.3\pm1.2$ \\
TWIST2 & HEFT & Ours & $47.8\pm1.9$ & $0.0\pm0.0$ & $0.0\pm0.0$ & $35.0\pm1.7$ & $24.4\pm2.5$ & $21.4\pm0.8$ \\
\bottomrule
\end{tabular}

\end{table}

\clearpage
\section{Compute Resources}
\label{app:compute-resources}

We run experiments on NVIDIA GeForce RTX 3080, RTX 3080 Ti, RTX 4090,
and H100 80GB GPUs. For the TTCO runs, wall-clock time
on an RTX 3080 Ti can be approximated by
\begin{equation}
    t(C,N)=\left(58.1+0.009961\frac{C}{N}\right)\ \text{s},
    \qquad R^2=0.890,
    \label{eq:runtime-fit}
\end{equation}
The $1/N$ factor reflects GPU parallelism across candidates. For large
datasets, motions share finite GPU capacity, so total computation $C$
remains relevant to compute-optimal allocation.

The runs in Secs.~4.1--4.3 used approximately 1,800, 380, and
1,700 RTX 3080 Ti-equivalent GPU hours, respectively. The runs in
Sec.~\ref{sec:loco-manipulation} used approximately 700 H100 GPU hours.

\end{document}